\documentclass{article} 
\usepackage{iclr2023_conference,times}

\usepackage{hyperref}
\usepackage{url}

\usepackage[utf8]{inputenc} 
\usepackage[T1]{fontenc}    
\usepackage{hyperref}       
\usepackage{url}            
\usepackage{booktabs}       
\usepackage{amsfonts}       
\usepackage{nicefrac}       
\usepackage{microtype}      
\usepackage{xcolor}         

\usepackage[pdftex]{graphicx}
\usepackage{caption}
\usepackage{subcaption}
\usepackage{hyperref}
\hypersetup{
    colorlinks=true,
    citecolor=black,
    linkcolor=black,
    urlcolor=blue,
    }
\usepackage{algorithm}
\usepackage{algpseudocode}
\usepackage{amsmath}
\DeclareMathOperator*{\argmax}{argmax} 

\usepackage{xspace}
\newcommand{\method}{$\mu$2Net+\xspace}

\usepackage{textgreek}
\usepackage[normalem]{ulem}

\mathchardef\mhyphen="2D

\title{A Continual Development Methodology for Large-scale  Multitask Dynamic ML Systems}

\author{%
  Andrea Gesmundo \\
  Google Research \\
  \texttt{agesmundo@google.com} \\
}

\begin{document}

\maketitle

\begin{abstract}
The traditional Machine Learning (ML) methodology requires to fragment the development and experimental process into disconnected iterations 
whose feedback is used to guide design or tuning choices.
This methodology has multiple efficiency and scalability disadvantages, such as leading to spend significant resources 
into the creation of multiple trial models that do not contribute to the final solution.
The presented work 
is based on the intuition that defining ML models as modular and extensible artefacts allows to introduce a novel ML development methodology enabling the integration of multiple design and evaluation iterations into the continuous enrichment of a single unbounded intelligent system.
We define a novel
method for the generation of dynamic multitask ML models as a sequence of extensions and generalizations.
We first analyze the capabilities of the proposed method
by using the standard ML empirical evaluation methodology.
Finally, we propose a novel continuous development methodology that allows to dynamically extend a pre-existing multitask large-scale ML system while analyzing the properties of the proposed method extensions.
This results in the generation of an 
ML model capable of jointly solving 124 image classification tasks achieving state of the art quality with improved size and compute cost.

\end{abstract}

\section{Introduction}

The traditional approach to the development of Machine Learning (ML) models assumes a fragmentation of the design and evaluation into multiple disconnected iterations causing inefficient use of available resources and engineering effort.
Alternative design choices are normally implemented and tested separately rather than being modularized and integrated in a search space whose exploration can be automated.
Furthermore each experimental iteration requires to reset the model's knowledge to a comparable initial state such as random initialization, thus preventing reuse or accumulation of knowledge across trials.
This fragmented development methodology is applied also to aspects such as hyperparameter tuning, as the training is usually repeated multiple times to compare different hyperparameter assignments even for a fixed model design.
This approach leads to spend a significant part of the available compute into the training of multiple models that provide no contributions to the final ML solution.
Furthermore, it relies on significant human intervention rather than aiming to automate any of the design or tuning exploration.

We propose and demonstrate a methodology that integrates the ML design and the experimental evaluation iterations into the continual expansion of an arbitrary pre-existing ML model.
The proposed ML methodology allows any design variation and model knowledge enrichment to be evaluated as an extension of the latest state of the dynamic ML system.
Successful extensions can be integrated as base for future  experimentation and further extensions.
This methodology enables the creation of dynamic multitask systems at a scale that would not be feasible to achieve with the constraint of running parallel repetitions from a neutral start state.

The two major contributions of the presented work are:
\begin{itemize}
    \item \textbf{A novel method for the generation of multitask dynamic systems}: We define a set of design and method extensions applicable to the $\mu$2Net method \citep{Gesmundo2022munet2},
    that provide improvement and generalization of different aspects such as: multi-factor model scoring,
    evolutionary agent capabilities and automated hyperparamter tuning (Section~\ref{section:method}).
    We refer to the novel method to as \method.
    The novel methods capabilities are first analyzed by using the traditional ML methodology on a smaller scale for feasibility (Section~\ref{section:preliminary}).
    \item \textbf{A novel ML development methodology integrating design and evaluation into the continual and unbounded extension of an ML system.}
    We  demonstrate empirically the effectiveness of the proposed methodology by incrementally extending the model generated by the experiments described in \cite{Gesmundo2022munet2}.
    The model is incrementally enriched as the \method improvements are introduced and as additional tasks are learned.
    This large scale continual experiments allows to analyze the effects of the method improvements (as done with the traditional methodology), while also continually extending the model into a large-scale multitask ML system capable of learning 124 image classification tasks with approximately state of the art quality and improved size and compute efficiency.
    (Section~\ref{section:experiments}).
\end{itemize}

\section{Related work}


Many methods have been proposed to achieve dynamic architecture extensions \citep{Chen2016Net2NetAL,Cai2018EfficientAS}, some also focusing on an unbounded stream of tasks \citep{Yoon2018LifelongLW}, some
achieving immunity from catastrophic forgetting  \citep{Rusu2016ProgressiveNN,Rosenfeld2020IncrementalLT} or optimizing for quality/cost trade-offs \citep{Tan2019MnasNetPN} (refer to Appendix~\ref{sec:ex-rel} for an extended survey).
Though, a methodology that integrates the continuous development of methods for dynamic architecture extensibility and knowledge enrichment of a multitask ML system jointly with the human design and experimentation process has not yet been proposed.

\cite{Berner2019Dota2W} describes a methodology that allows for the continued variation of model architecture and training method while preserving its knowledge.
Although, the dynamic architecture extension capabilities are not provided by the definition of higher level abstractions 
but rather require \emph{surgery} operations requiring custom human intervention.
\cite{Berner2019Dota2W} tackles a single high-complexity task, and the ability of exploring architecture and method variations without resetting the knowledge is deemed a critical accelerator in such context.
Advancements in ML models dynamicity, extensibility and reusability can contribute to accelerate progress toward further high-complexity achievements such as "Artificial General Intelligence" (AGI).

The ability of jointly solve a large amount of tasks is another theme that is commonly associated with progress toward AGI.
Advancements in scaling language models \citep{Brown2020LanguageMA,Thoppilan2022LaMDALM} allowed to achieve discourse and reasoning capabilities that can be applied to new tasks without requiring additional training.
Recent work aims to extend these achievements beyond text modality by defining static architectures for an extended subset of modalities \citep{Alayrac2022FlamingoAV,Reed2022AGA}.
These are few examples of the ML models contributing to the line of research achieving incremental milestone toward AGI. Though, each model is trained from scratch with considerable resources consumption.
The introduction of abstractions allowing to modularize, dynamically extend and reuse these large models may contribute to accelerate the rate of innovation. 

\section{Method}
\label{section:method}

The proposed method extends $\mu$2Net \citep{Gesmundo2022munet2} with the intent of generalizing and improving its capabilities.
In this section we detail the four method extensions introduced with \method:
1) multi factor scoring function, 2) improved hyperparameters search space, 3) additional mutation actions, 4) learned $\mu(\cdot)$ function.
Refer to \cite{Gesmundo2022munet2} for a detailed definition of the $\mu$2Net method being extended.

The first method extension consists of the introduction of a new \textbf{multi factor scoring function} that integrates additional size and compute penalty factors by employing an exponential decay penalty:
\begin{equation}
\textstyle  
score(m) = q(m) * s ^{\left( \frac{ \#accounted\mhyphen params(m) }{ P } \right)} * s ^ {\left( \frac{ \#flops(m) }{ F } \right)}
\end{equation}
$q(m)$ denotes the quality metric achieved by model $m$ on the validation set.
$s\in\ ]0, 1]$ is the cost penalties scaling factor.
$\# flops(m)$ denotes the amount of floating point operations required by model $m$ at inference time,
and $\# accounted\mhyphen params(m)$ is the sum of parameters used by model $m$,
dividing each parameter count by the number of models 
sharing its use:
\begin{equation}
\textstyle  
\#accounted\mhyphen params(m) = \sum_{p\in \mathcal{P}(m)} \frac{1}{\#models(p)+1} 
\end{equation}
$\mathcal{P}(m)$ denotes the set of all parameters used by $m$.
$\#models(p)$ denotes the count of models for tasks different from the task $m$ is trained on, that are currently sharing the use of parameter $p$.
Intuitively, if $s\!=\!0.99$, then a relative reduction of $1\%$ of the quality metric is applied if a tasks is accounted for P parameters or requires F flops.

\begin{table}[t]
  \caption{
Hyperparameters search space.
Sequences of valid values for each automatically tunable hyperparameter.
Bold vales are defaults.
Underlined values are additional values or changes of defaults in comparison to the search space defined by $\mu$2Net.
}
  \label{table:hps}
  \centering
  \begin{tabular}{l}
  \toprule
\multicolumn{1}{c}{\emph{Optimizer hyperparameters}} \\
Learning rate $\in$ [0.0001, 0.0002, 0.0005, 0.001, 0.002, 0.005, \textbf{0.01}, 0.02, 0.05, 0.1, 0.2, 0.5] \\
Learning rate schedule warm up ratio $\in$ [\underline{0}, 0.01, 0.02, 0.05, \textbf{0.1}, 0.2, 0.3] \\
Momentum $\in$ [0.5, 0.6, 0.7, \underline{0.75}, 0.8, 0.85, \textbf{0.9}, 0.95, 0.98, 0.99] \\
Nesterov update $\in$ [\textbf{False}, True] \\
\midrule
\multicolumn{1}{c}{\emph{Data Preprocessing hyperparameters}} \\
Cropped area range min $\in$ [\underline{0.05}, 0.5, \underline{0.95}, \underline{\textbf{1.0}}] \\
Cropped aspect ratio range min $\in$ [0.5, 0.75, \underline{\textbf{1.0}}] \\
Flip left/right $\in$ [\underline{\textbf{False}}, True] \\
Brightness delta $\in$ [\textbf{0.0}, 0.01, 0.02, 0.05, 0.1, 0.2] \\
Contrast delta $\in$ [\textbf{0.0}, 0.01, 0.02, 0.05, 0.1, 0.2] \\
Saturation delta $\in$ [\textbf{0.0}, 0.01, 0.02, 0.05, 0.1, 0.2] \\
Hue delta $\in$ [\textbf{0.0}, 0.01, 0.02, 0.05, 0.1, 0.2] \\
Image quality delta $\in$ [\textbf{0.0}, \underline{0.01}, \underline{0.02}, \underline{0.05}, \underline{0.1}, \underline{0.2}] \\
Image resolution pixels $\in$ [\underline{224}, \textbf{384}] \\
\bottomrule
  \end{tabular}
\vspace{-12pt}
\end{table}

\begin{figure}
  \centering
  \includegraphics[width=1.0\linewidth]{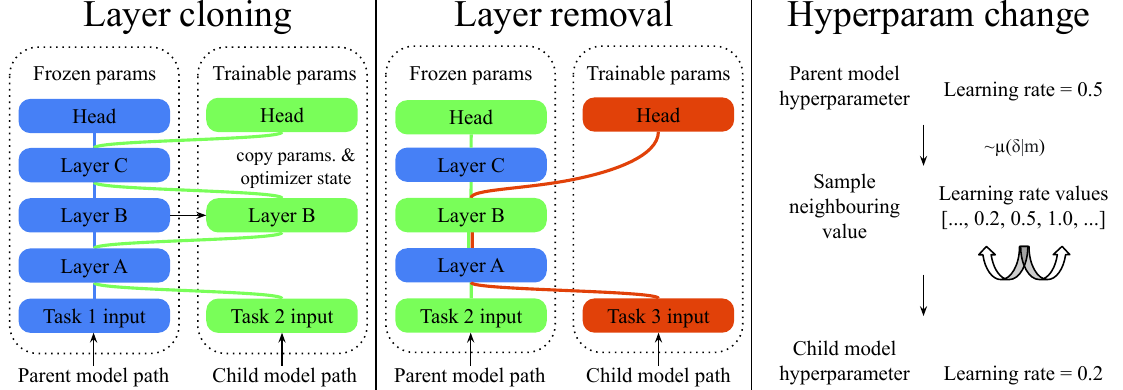}
  \caption{Graphical representation of the three types of mutations defined by the proposed method.
New models are generated by applying a sampled subset of possible mutations.
}
\label{fig:mut}
\end{figure}

The base $\mu$2Net method allows for 2 types of \textbf{mutation actions}: 1) any layer from the parent model can be cloned to create a trainable copy carrying its parameters and optimizer state (e.g. momentum statistics), any other layer is shared in a frozen state,
2) any hyperparameter value can be changed to one of the neighbouring values in the sequence of valid values defined by the search space (see Table~\ref{table:hps}).
\method includes a third mutation type, layer removal: the top transformer layer of the parent model can be removed from the sequence of layers defining the child models (see Figure~\ref{fig:mut}).
The addition of this mutation type allows to sample models with a number of transformer layers that is smaller than the number of layers of the root model.
This allows to generate models using less parameters and compute, thus introducing an additional degree of freedom to optimize for the introduced cost factors.

The hyperparameters \textbf{search space extension} includes different types of additions (see Table~\ref{table:hps}).
All the models generated by $\mu$2Net are constrained to use 384 pixels image resolution. This is the resolution recommended for ViT fine-tuning \citep{Steiner2021HowTT}. Although the ViT model we are using as a root model has been pretrained on 224 resolution.
We extend the search space to include the option of using 224 resolution.
This provides an additional option to generate models requiring less compute, as one of the objectives of the empirical study is to analyze the quality/compute trade-off.
The options for image preprocessing are extended by adding the possibility to alter the image quality.
This is a commonly used technique that was not included in the recommended fine-tuning setup \citep{Steiner2021HowTT} and therefore was not included in $\mu$2Net search space.
Though, we find that this new preprocessing function is learned to be activated for a significant portion of the generated models configuration even if set as inactive by default (see Figure~\ref{fig:dists}).
Rarely used values have been removed.
\cite{Steiner2021HowTT} recommends a mild preprocessing configuration for ViT fine-tuning.
The preprocessing hyperparameter default values for $\mu$2Net search space were set to match the recommended configuration.
Instead, \method set the default values to disable all the preprocessing.
This change is based on the observation done during preliminary experiments that preprocessing can moderately benefit most of the tasks, but it can have critical negative impact for a subset of tasks. 
New values have been added to the valid values ranges of some hyperparameters based on observations done over the distributions learned in previous experiments such as those reported in \cite{Gesmundo2022munet2}. For example, the most common selected value for the "learning rate schedule warm up ration" was the edge value of $0.01$, thus the range has been extended by adding the value $0$, that allows to disable the learning rate warm up.
Also notice that $\mu$ has been removed from the search space as a hyperparameter since it is now defined as a learnable function conditioned on the mutation type, as described in the following paragraph.

In \citet{Gesmundo2022munet1} the mutation sampling is parameterized by a global mutation probability constant $\mu$.
\citet{Gesmundo2022munet2} extends the approach by defining $\mu$ as a per model hyperparamter tuned by the evolutionary method, thus enabling the selection of distinct mutation probabilities for each model and adapt the their value through time.
\method allows for further per action customization of the learned mutation probabilities by defining a $\boldsymbol\mu$ \textbf{function}, $\mu(\delta|m)$, for each model $m$.
This function maps each mutation action, $\delta$, to the probability of applying the mutation action.
This extension enables the selection of a distinct mutation probability for each possible mutation action.
In practice, the $\mu$ function is implemented as look-up table mapping every possible mutation to a distinct mutation probability value.
The values of the lookup table are inherited and mutated across generations following the same incremental value mutation logic applied to the hyperparameters.
All the values in the lookup table are initialized to $0.2$, and the sequence of valid mutation probability values is defined as the multiples of $0.02$
in the range $[0.02, 0.3]$.

\section{Parallel training comparison}
\label{section:preliminary}

\begin{figure}[t]
\centering
\text{Multitask Character Classification Benchmark}\par\medskip
\begin{minipage}{.5\textwidth}
  \centering
  \includegraphics[width=1.0\linewidth]{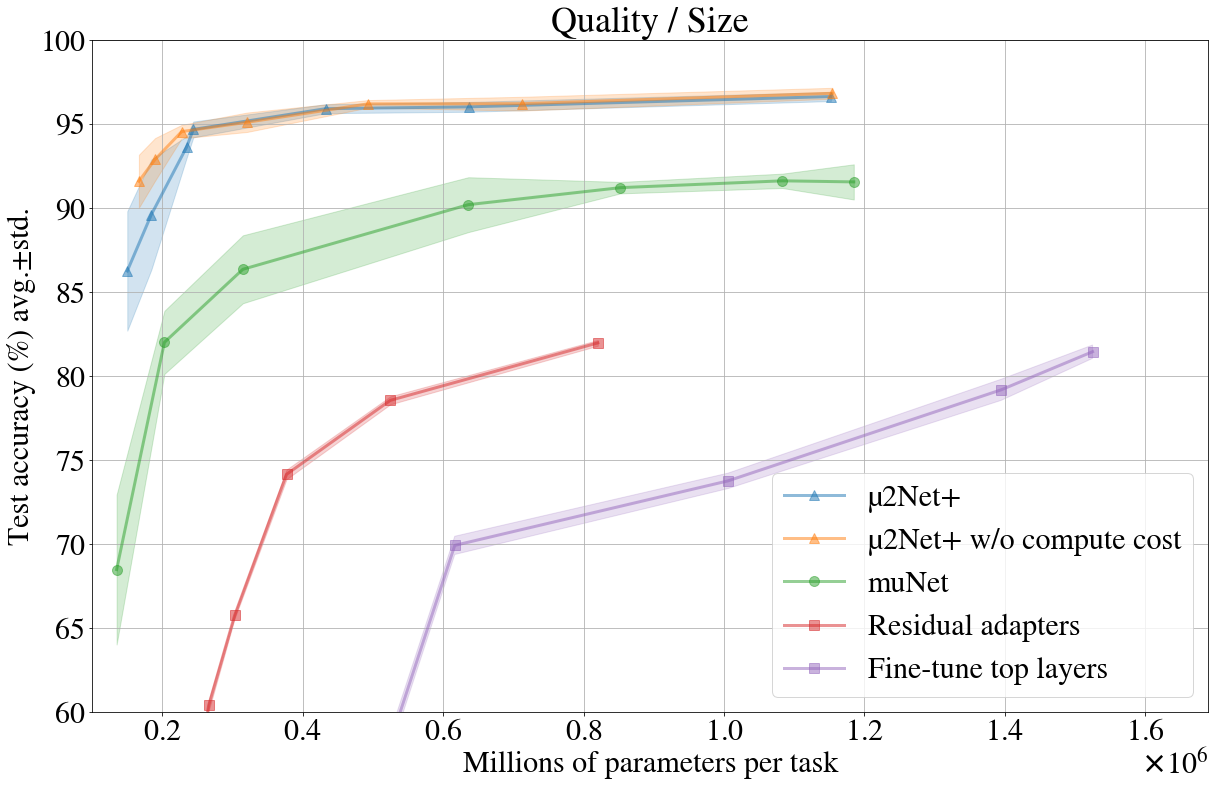}
\end{minipage}%
\begin{minipage}{.5\textwidth}
  \centering
  \includegraphics[width=1.0\linewidth]{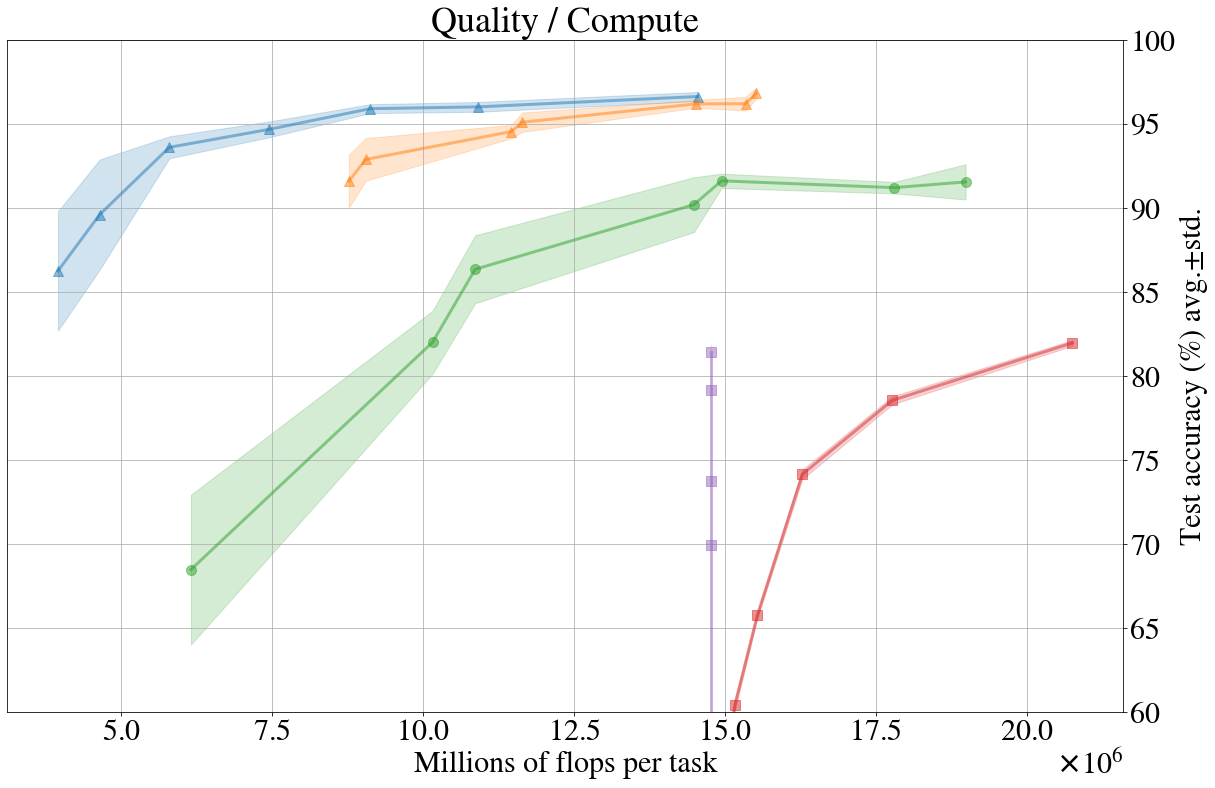}
\end{minipage}
\par\medskip
\text{Visual Domain Decathlon Benchmark}\par\medskip
\begin{minipage}{.5\textwidth}
  \centering
  \includegraphics[width=1.0\linewidth]{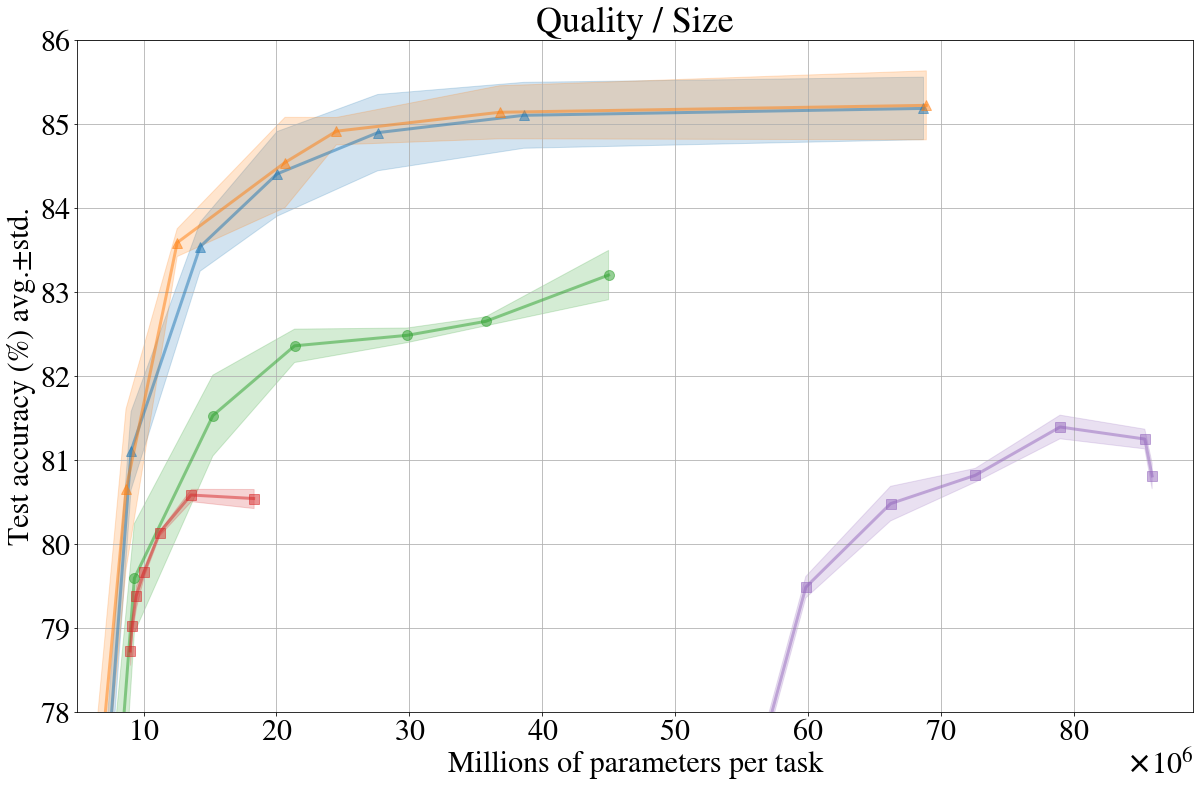}
\end{minipage}%
\begin{minipage}{.5\textwidth}
  \centering
  \includegraphics[width=1.0\linewidth]{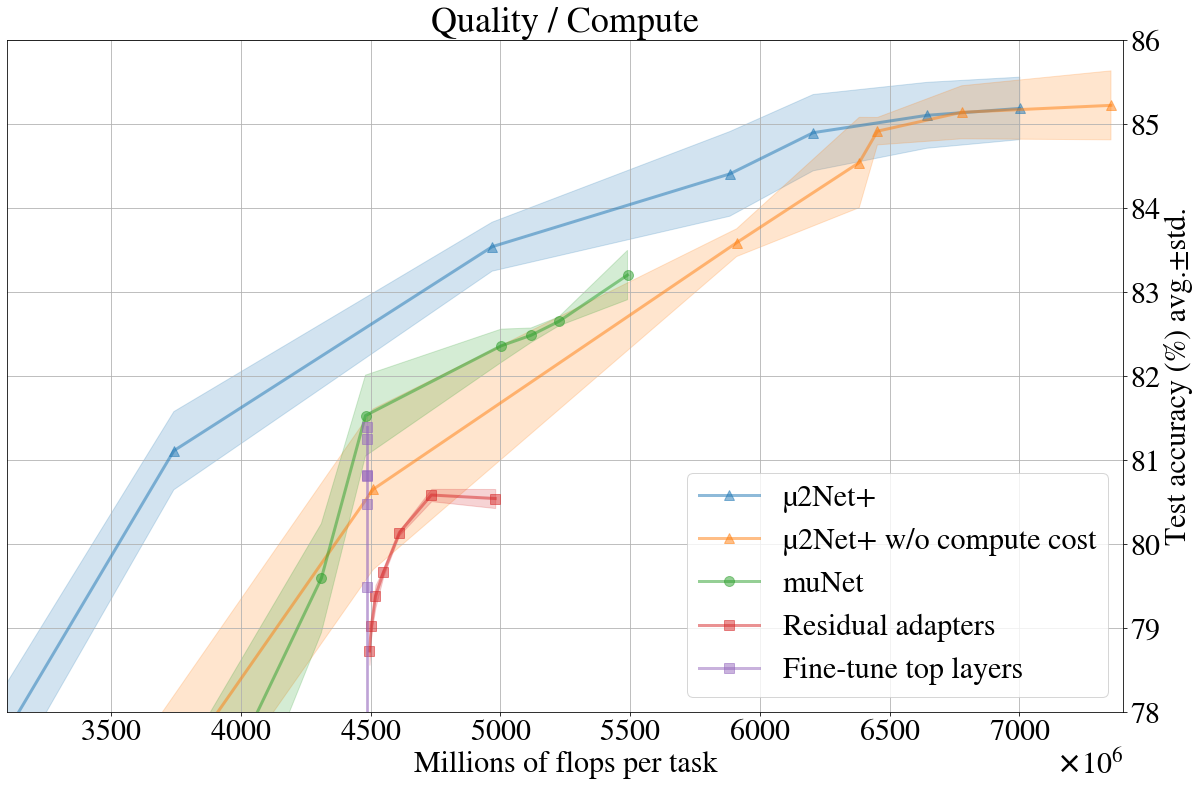}
\end{minipage}
\caption{Pareto frontiers achieved in the quality/size and quality/compute spectrum by \method and baseline methods on the Multitask Character Classification Benchmark and the Visual Domain Decathlon Benchmark (see Section~\ref{section:preliminary}).}
\label{fig:curves}
\end{figure}

In this section we use the standard ML methodology to 
compare different method by running parallel replicated experiments each using comparable configuration and amount of compute.
In particular we focus on quality/cost trade-off comparison.
Thus, we compare against common baseline methods that allow to explore the Pareto frontier of such trade-offs.
We compare against the muNet method \citep{Gesmundo2022munet1}, and against standard transfer learning techniques such as fine-tuning and residual adapters that provide a cover of the considered
trade-off spectrum.
As the proposed method introduce the joint optimization of quality, size and compute,
we consider two trade-offs: quality/size and quality/compute.

For comparability, the experiment setup and benchmarks are equivalent to those defined in \cite{Gesmundo2022munet1} and are summarized in this paragraph.
To test the generality of the results, we repeat the experiments on 2 benchmarks:
1) the Multitask Character Classification Benchmark (MCCB), composed by 8 character classification tasks,
2) the Visual Domain Decathlon Benchmark (VDDB), composed by the 10 tasks of the Visual Domain Decathlon challenge. The VDDB is more computational intensive and tasks have been selected to represent a diverse set of domains (see Table~\ref{table:datasets2} for datasets details and references).
Experiments are run on TPUv3 with 8 parallel cores. 
Each experiment is repeated multiple times to measure variance: 5 times for MCCB and 3 times on VDDB (the lower number of repetitions is due to the higher experiment cost).
Every experiment is allocated an equivalent training budget in terms of training epochs per task. Epochs are capped to 51200 training samples for all datasets to smooth the distribution of compute over the set of tasks.
For muNet and \method models different trade-offs are achieved by setting the scale factor, $s$, to different values in the set: $\{0.02, 0.3, 0.7, 0.9, 0.95, 0.98, 1\}$.
For the Residual adapters baselines, different trade-offs are achieved by changing the inner dimension of each adapter to different values in the set: $\{8 ,16, 32, 64, 128, 256, 512\}$ and all the non residual adapter layers are frozen and shared across models for different tasks.
While, for the Fine-tune top layers baseline, different trade-offs are achieved by freezing an increasing number of bottom layers, and again all the frozen layers are shared across multiple tasks to decrease the average number of parameters per task.

The only difference between the experiment configuration of the baseline muNet method and the proposed \method method is in how the equivalent training budget is distributed.
For \method, the training budget is distributed to perform more but shorter task iterations.
The motivation for this change is to bring the ratio between number of task iterations and length of each iteration closer to what would be representative of the setup for a long term continual learning experiment, that require a significantly higher number of task iterations than the 2 that were originally allocated to muNet experiments.
Thus we change the distribution to increase the similarity with the setup used by the large scale experiments reported in \cite{Gesmundo2022munet2} and their extension with the novel methodology reported in Section~\ref{section:experiments}.
These experiments perform 4 training epochs per generation, 4 generations per task iteration and allow for an unbounded number of task iterations.
In details, the training budget distribution change consists of: for the MCCB, muNet is configured to perform 2 task iterations, 8 generations for each task iteration and each model is trained for 5 epochs, while \method performs 5 task iterations, 4 generations for each tasks, and 4 training epochs per model, totalling the same amount of training epochs per task.
While, for the VDDB, muNet is configured to perform 2 task iterations, 8 generations for each task iteration and each model is trained for 30 epochs, while \method performs 30 task iterations, 4 generations for each tasks, and 4 training epochs per model.

The \method size cost factor is configured with parameters equivalent to those used for muNet: $P\!=\!1.48$M for MCCB and $P\!=\!85.6$M for VDDB proportionally to the number of parameters in the root model.
While for the compute cost factor we set: $F\!=\!15$M for MCCB and $F\!=\!100$M for VDDB proportionally to the inference flops required the root model with an image size of the higher end of the allowed range.

The results are represented graphically in Figure~\ref{fig:curves}.
The overall comparison shows that the proposed method outperforms all the baseline methods in both benchmarks across both the quality/size and quality/compute spectrum.
We also compare with a version of \method without compute cost factor.
In the quality/size spectrum we observe that the addition of the compute cost factor does not alter significantly the achieved Pareto frontier.
While, we notice a significant gain in the quality/compute trade-off.
For example, notice that \method without compute cost factor does not outperform muNet in lower end of the quality/compute spectrum for the VDDB experiments.
The metrics of the configurations achieving the best quality for each method are reported in Table~\ref{table:bestmetrics}.

It is also interesting to notice that the curves representing the Pareto frontier achieved by Fine-tune top layers becomes a vertical line in the quality/compute spectrum, because the inference compute is independent from the number of layers that are frozen and shared.
Also notice that the Residual adapters Pareto frontier outperforms the Fine-tuning one in the quality/size domain.
That is expected since Residual adapters are usually employed as a parameter-efficient transfer-learning technique. 
Although, the order is inverted in the quality/compute spectrum, since Residual adapters add compute cost on top of the root model.

\begin{table*}[t]
\begin{center}
\caption{
Methods comparison on the MCCB and VDDB task-sets.
The metrics of the configuration achieving the best quality are reported for each method.
Quality is measured as the test accuracy averaged across tasks.
The table reports the max, average and standard deviation of the quality achieved by the experiment repetitions.
Model size is measured as the average number of parameters per tasks. Compute is measured as average inference flops.
}
\label{table:bestmetrics}
\begin{tabular}{lcccc}
\toprule
 & \multicolumn{2}{c}{Test Acc. \%} & Params/task & Flops/task\\
                     Model &    Max &   Avg.$\pm$Std. & ($\times10^6$) & ($\times10^6$) \\
\midrule
\multicolumn{5}{c}{\emph{Multitask Character Classification Benchmark}} \\
           Full fine-tuning &  82.20 &  81.46$\pm$0.40 &                1.53 &  14.77 \\
 Residual adapters dim=512 &  82.28 &  81.98$\pm$0.17 &                0.82 &  20.76 \\
                     muNet &  92.98 &  91.41$\pm$1.06 &                1.20 &  16.64 \\
                $\mu$2Net+ &  97.05 &  96.65$\pm$0.27 &                1.15 &  14.55 \\

\midrule
\multicolumn{5}{c}{\emph{Visual Domain Decathlon Benchmark}} \\
          Fine-tune above 1st tr. layer &  81.51 &  81.39$\pm$0.14 &               78.98 &  4485
          \\
 Residual adapters dim=256 &  80.67 &  80.58$\pm$0.07 &               13.56 &  4734
 \\
                     muNet &  83.58 &  83.20$\pm$0.29 &               45.00 &  5490
                     \\
                $\mu$2Net+ &  85.56 &  85.19$\pm$0.37 &               68.70 &  7001
                \\
\bottomrule
\end{tabular}

\end{center}
\end{table*}

\section{Continual model extension}

\label{section:experiments}

In this section we describe and demonstrate the proposed continual development methodology that aims to analyze the properties of method and knowledge augmentations through a sequence of experiments while dynamically extending a single large-scale ML model.
With this methodology, we aim to characterize and quantify the effects of the introduced extensions by measuring changes 
thought time, rather than comparing the metrics achieved by parallel runs of method variations (as done with the methodology used in Section~\ref{section:preliminary}).
The proposed continual development approach has the following advantages:
\begin{enumerate}
\item allows to focus all the available compute on the enrichment of a single ML system, thus increasing the accessibility of larger scale experiments,
\item lifts the requirement of training multiple model variations that do not contribute to the final solution for the tasks at hand,
\item enables the unbounded extension of dynamic large-scale multitask systems as the one considered in this section, since it would not be feasible at this scale to 
train 
multiple randomly initialized copies of such systems to evaluate each method variation that can be interesting to analyze.
\end{enumerate}

The following experiments also demonstrate how the proposed \method method can be applied to extend a pre-trained multitask system generated with a different method.
The large-scale multitask system generated by the experiments described in \citep{Gesmundo2022munet2} is used as the \emph{starting multitask system}. 
This starting multitask system can already solve a set of 69 image classification tasks.
With \emph{task-set A} we refer to that first set of 69 tasks.
With \emph{task-set B} we refer to an additional set of 55 image classification tasks that will be introduced in following experiments.

The continual extension of the system is described as a sequence of experiment segments.
Each 
segment introduces extensions to the method or task-set.
The the effects on the system in terms of quality, efficiency and capabilities are measured through additional task iterations.

\begin{figure}[t]
  \centering
  \includegraphics[width=1.0\linewidth]{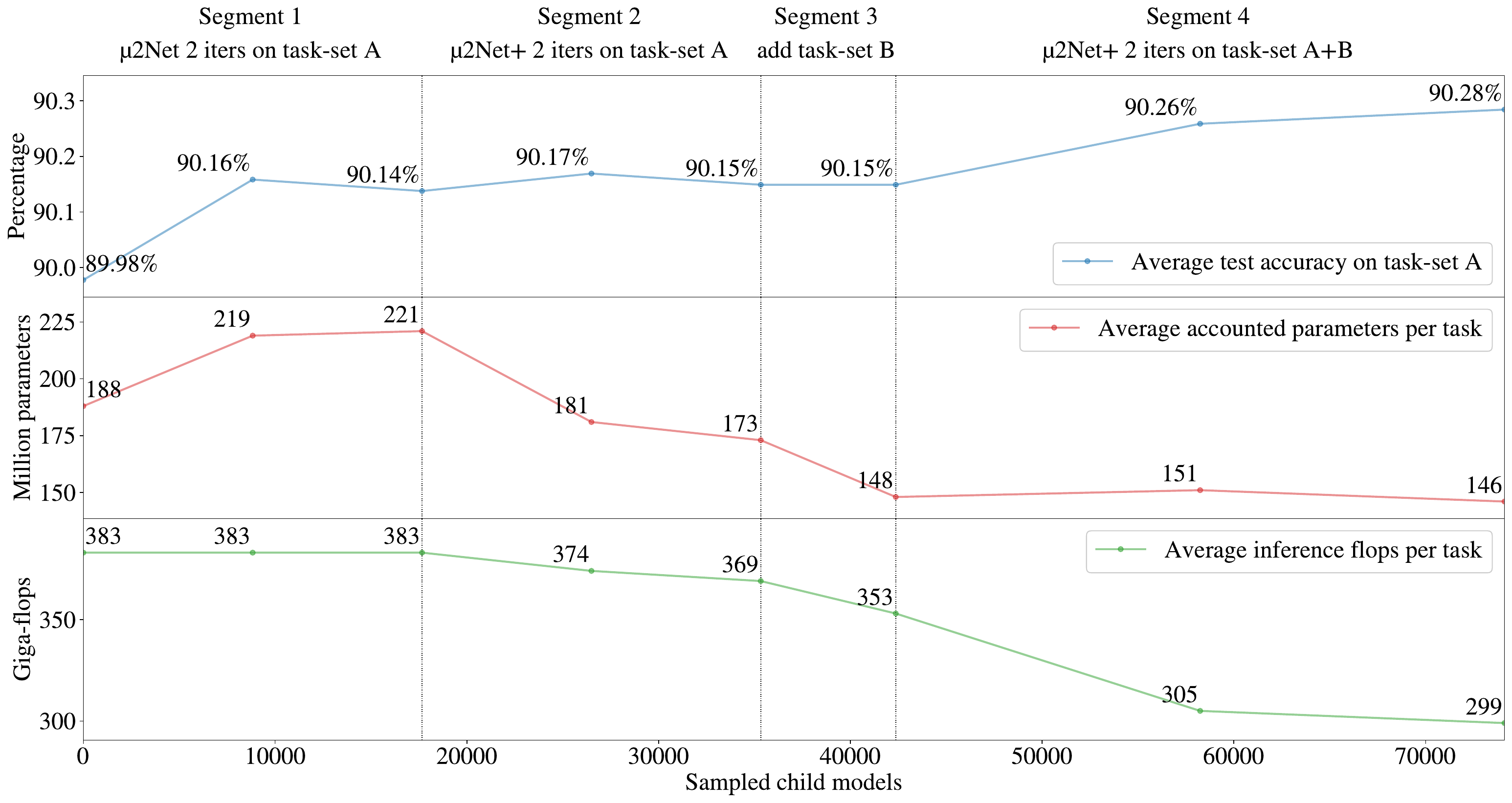}
  \caption{
  Displays variations through the sequence of experiment segments 
  of 3 reference metrics:
  1)~Average test accuracy on the task-set A, defined as the initial 69 image classification tasks introduced in \cite{Gesmundo2022munet2}, notice that the reported test accuracy is computed on a test set that does not overlap with the training set and validation set (whose validation accuracy is used for the reward computation) to avoid overfitting and selection bias.
  2)~Accounted parameters averaged across all the models composing the multitask system at the time of measure (not limited to task-set A). This measure corresponds to the size cost factor included in $\mu$2Net+ reward function (see Section~\ref{section:method}).
  3)~Compute cost measured as flops required to produce inference for one input sample, also averaged across all the models composing the multitask system at the time of measure (not limited to task-set A). This measure corresponds to the compute cost factor included in $\mu$2Net+ reward function (see Section~\ref{section:method}).
  }
  \label{figure:continual}
\end{figure}

\subsection{Experiment segment 1: $\mu$2Net convergence}

The first experiment segment applies the $\mu$2Net method as defined in \citep{Gesmundo2022munet2} for an additional number of tasks iterations with the aim of achieving a sufficient system stability and convergence on task-set A.
Achieving convergence with the baseline method before transitioning to an extended version allows to attribute the effects observed the following experiment segments to the corresponding method extensions, as the method extensions are compared sequentially rather with standard parallel experiments.

We observe that 2 additional task iterations are required for $\mu$2Net to achieve sufficient convergence on task-set A (see Figure~\ref{figure:continual}).
During the first task iteration, the average test accuracy increases by 0.18\%.
While during the second task iteration, the test accuracy does not improve.
The trend of the "average accounted parameters per task" metric is similar, as the variation is significantly smaller during the second iteration.
Notice that all the models generated by the $\mu$2Net method have a constant compute cost, that matches the compute cost of the used ViT Large root model with input image resolution of 384$\times$384 pixels.
This is the case since the $\mu$2Net method does not provide mutation actions that can alter the compute cost of the generated model, such as adding/removing transformer layers or changing the image resolution. 

\subsection{Experiment segment 2: \method introduction}

The second experiment segment introduces the proposed \method method.
To produce mild size and compute cost penalties, the scoring function parameters $P$ and $F$ are set respectively to $\sim\!10$ times the accounted parameters and inference flops averaged across all the models that are part of the system at the beginning of this experiment segment: $P\!=2.2\!\times\!\!10^{9}$ and $F\!=\!3.8\!\times\!10^{12}$.
$s$ is set to $0.99$.
All the common evolutionary agents meta parameters, such as those determining the compute utilized per model training and exploratory budget per task iteration, are unchanged for comparability \citep{Gesmundo2022munet2}. 

Two additional task iterations on the task-set A are performed. 
The metrics reported in Figure~\ref{figure:continual} show that
the introduction of the novel method leads the newly generated models to share more parameters and be more computationally efficient.
Despite this, the average test quality does not deteriorate.
Overall resulting in a 3.6\% reduction in compute and 21.7\% reduction in size with no loss in quality.

This efficiency improvement is enabled by:
1) the generation of cheaper models is enabled by the introduction of mutation actions allowing to reduce the number of transformer layers and the extension of the hyperparameter search space allowing to chose smaller input image resolution,  
2) cheaper models selection is encouraged by the introduction of the size and compute cost factors. 
3) the learning process of the $\mu(\cdot)$ function increases the probability of sampling cheaper models by customizing the probability of occurrence of different mutation types.
For example, the probability of performing a layer cloning action decreases from 21.4\% to 17.4\%, thus decreasing the expected accounted parameter cost of newly sampled models.
Furthermore, the learned $\mu(\cdot)$ functions appear to assign lower cloning probability to lower layers in average (see Figure~\ref{fig:clone-prob}).

\subsection{Experiment segment 3: task-set B introduction}

The third experiment segment consists in extending the number of jointly solved tasks from 69 to the 124.
This is achieved by performing one task iteration on the additional $55$ tasks of task-set B to introduce them into the system.
These additional image classification tasks are also publicly distributed via the \href{https://www.tensorflow.org/datasets/catalog/overview}{Tensorflow Datatasets catalog}.

The average test accuracy achieved on task-set B after one task iteration is 90.66\%.
If we compare to the accuracy achieved at convergence in the next experiment segment (90.78\%), we can observe that the system is able to achieve close to converge quality with only 
one task iteration.
While, it took more task iterations to achieve the same relative delta to convergence quality on task-set A: at least 4 task iterations (including the task iteration described in \cite{Gesmundo2022munet2}).

This observation can be an empirical evidence supporting the following hypothesis: \emph{The expected convergence time for a new task is inversely proportional to the amount of knowledge embedded in the system}.
This hypothesis is based on the intuition that a new task can be learned faster if a model or sub-set or parameters with as relevant knowledge as possible is available to be used or fine-tuned.
And the probability for a new task of fining relevant knowledge within the system increases as the system learns more tasks.
This hypothesis can be also supported by the sharp drop in "average accounted parameters per task": 14.5\% relative reduction during this iteration (see Figure~\ref{figure:continual}), as the new models added to the system seem to take advantage of layers and knowledge sharing with a higher degree.

The average test quality measured on task-set A remains unchanged since its tasks are excluded from this iteration.
The average inference compute per task appear to continue to decrease: $4.3\%$ relative reduction in this iteration.

\subsection{Experiment segment 4: $\mu$2Net+ convergence}

To conclude we perform 2 additional task iterations on the union of task-sets A and B.

Interestingly, the average test quality measured on task-set A improves significantly after achieving no significant changes for 3 consecutive task-iterations (see Figure~\ref{figure:continual}).
With respect task-set A, the only variation introduced in this experiment segment is the possibility of accessing the new knowledge introduced by learning to solve the additional 55 tasks of task-set B.
We measure that, during this experiment segment, 15 tasks of task-set A find an better scoring model by transferring knowledge from at least one new task of task-set B, and 9
of these improved models transfer knowledge from 2 or more task-set B tasks.

The average inference flops also further improves.
Computationally cheaper models are rendered possible by the introduction of the possibility to remove transformer layers and use smaller input image resolutions.
Before the introduction of the proposed method all the models used 384 pixels input image resolution and 24 transformer layers as the ViT Large root model.
Since the introduction of \method, 30.4\% of the 124 selected models use 224 image resolution, and 27.2\% use 23 transformer layers, 1.6\% use 22 and 2.4\% use 21 transformer layers (see Figure~\ref{fig:dists}).

\section{Conclusion}
\label{section:conclusions}

The reported series of system and method extensions demonstrates the ability of the proposed ML development methodology of integrating the continual design and evaluation process into the unbounded extension of large-scale multitask ML systems. 
The proposed continual development methodology not only allows for the continual addition of knowledge and tasks into a dynamic multitask system, but also allows for dynamic extensions of the search space defining the architectures and configurations that are possible to explore and furthermore allows to iteratively extend the method exploring such search space while also adapting through time the guiding reward function to adjust according to possibly changing requirements.

Overall, the methodology and empirical results demonstrated in by this work are encouraging signs that a highly automated and
incrementally extensible ML system for handling thousands or millions of tasks with improved quality and efficiency is achievable.
Future work can continue to build toward a system that can acquire further capabilities and knowledge across multiple modalities.

\clearpage



\bibliography{iclr2023_conference}
\bibliographystyle{iclr2023_conference}


\clearpage
\appendix

\section{Extended related work survey}
\label{sec:ex-rel}
The proposed method is designed to learn an unbounded number of tasks in a continual learning fashion.
In such contexts it aims to learn each task with higher quality and efficiency by automating and optimizing the knowledge transfer among any subset of tasks that can provide useful knowledge to one another.
The proposed model is designed to be immune from common multitask learning pitfalls: catastrophic forgetting, gradients interference, negative transfer. 
Cross-task \textbf{transfer-learning}
has gained popularity, especially through transfer learning from a model
pre-trained on a large amount of data for one or a few general tasks,
and then fine-tuned on a small amount of data for a related downstream task.
This approach has been shown to be very effective in a wide variety of problems
across many modalities, including
language \citep{Devlin2019BERTPO,Raffel2020ExploringTL} and vision \citep{Dosovitskiy2021AnII,He2016DeepRL}.
The success of transfer-learning applications hinges on adequate prior knowledge selection to avoid typical \textbf{negative transfer} pitfalls \citep{Rosenstein2005ToTO,Wang2019CharacterizingAA}.
Common solutions rely on data or model selection techniques,
often putting emphasis on the efficiency of the exploration
\citep{Zhang2020ASO,Mensink2021FactorsOI}, also method aiming to automate knowledge selection at a layer level have been proposed \citet{Sun2020AdaShareLW}.
Transfer learning capabilities are critical for \textbf{multitask models}.
ML models trained jointly on multiple tasks 
can be affected by \textbf{gradients interference} if any subset of parameters receive gradients jointly from multiple sources \citep{Chen2018GradNormGN,Yu2020GradientSF}, and by \textbf{catastrophic forgetting} of prior knowledge as new tasks are learned \citep{McCloskey1989CatastrophicII,French1999CatastrophicFI}.
These knowledge loss problems can be alleviated with weighted combination of tasks \citep{Liu2019LossBalancedTW,Sun2020ERNIE2A} and gradient transformation methods \citep{Chen2018GradNormGN,Sener2018MultiTaskLA,Kendall2018MultitaskLU}.
Stronger guarantees are provided by methods that compartmentalize task specific knowledge in dedicated parameter subsets \citep{Rebuffi2017LearningMV,Houlsby2019ParameterEfficientTL,Rusu2016ProgressiveNN,Rosenfeld2020IncrementalLT}.
Addressing catastrophic forgetting and identifying what subset of parameters/knowledge that is beneficial to share with each task is also critical for \textbf{continual learning} or life long learning methods
\citep{McCloskey1989CatastrophicII,French1999CatastrophicFI,Ramesh2022ModelZA}.

The proposed method relies on an evolutionary approach to jointly search the spaces of models architectures, hyperparameters, and prior knowledge selection while optimizing for an possibly multi-factor non-differetiable reward function.
The automation of \textbf{hyperparameter tuning} has been commonly addressed with Bayesian optimization \citep{Srinivas2010GaussianPO,Bergstra2011AlgorithmsFH,Snoek2012PracticalBO},
evolutionary methods have also been explored for this purpose
\citep{Jaderberg2017PopulationBT,Zhang2011EvolutionaryCM}.
Hyperparameters tuning can be considered related to the \textbf{neural architecture search} (NAS), as architectures can be defined by the selection of a sequence
of architectural hyperparameters.
Initially, NAS methods have been based on reinforcement learning techniques
\citep{Zoph2017NeuralAS} but also sample efficient evolutionary approaches have been proposed \citep{Real2019RegularizedEF,Maziarz2018EvolutionaryNeuralHA}.
Parameter-sharing based NAS methods aim to reduce the typically high training cost \citep{Pham2018EfficientNA,Liu2019DARTSDA,Kokiopoulou2019FastTA}.
Optimization for multi-factor quality/cost trade-offs have been explored \citep{Tan2019MnasNetPN}.

The proposed method is capable to dynamically extend the system, adding capacity or novel structures in an unconstrained fashion.
A few methods have been proposed to achieve \textbf{dynamic architecture extensions} \citep{Chen2016Net2NetAL,Cai2018EfficientAS}, some also focusing on an unbounded stream of tasks \citep{Yoon2018LifelongLW,Yao2020OnlineSM}, or achieving immunity from catastrophic forgetting  \citep{Rusu2016ProgressiveNN,Li2018LearningWF,Li2019LearnTG,Rosenfeld2020IncrementalLT}.

The proposed method is sparsely activated, thus the unbounded growth of knowledge and parameters is decoupled from the growth of computational cost.
The growth in capabilities of state of the art models often requires growth in terms of trainable parameters \citep{Kaplan2020ScalingLF}.
\textbf{Sparse activation} techniques at sub-layer level \citep{Shazeer2017OutrageouslyLN,Du2021GLaMES} or network route level \citep{Fernando2017PathNetEC} allow to decouple model size growth from compute cost.
This is achieved by integrating 
a \textbf{routing technique} that selects the appropriate subset of parameters storing the most relevant knowledge for each task, sample or token/patch.

The ability of jointly solve a \textbf{large amount of tasks} is commonly associated with progress toward Artificial General Intelligence (AGI).
Advancements in scaling language models \citep{Brown2020LanguageMA,Thoppilan2022LaMDALM} allowed to achieve novel discourse, reasoning and zero/few shot learning capabilities that can be applied to new tasks without/minimal additional training.
Recent work aims to extend these achievements beyond text modality by defining static architectures for an extended subset of modalities \citep{Alayrac2022FlamingoAV,Reed2022AGA}.
These are few examples of the ML models contributing to the line of research achieving incremental milestone toward AGI.
Though, each model is trained from scratch with considerable resources consumption.
The introduction of abstractions allowing to modularize, dynamically extend and reuse these large models may contribute to accelerate the rate of innovation. 







\section{Experiments details}
\label{section:repro}
All the experiments reported in this paper can be reproduced by using the following public resources:
\begin{itemize}

\item The published code of the proposed \method:
\href{https://github.com/google-research/google-research/tree/master/muNet}{https://github.com/google-research/google-research/tree/master/muNet}

\item $\mu$2Net and the large scale model checkpoint: \href{https://github.com/google-research/google-research/tree/master/muNet}{https://github.com/google-research/google-research/tree/master/muNet}

    \item 
The ViT model definition and checkpoints published by \citet{Steiner2021HowTT}.
These resources are available at  \href{https://github.com/google-research/vision_transformer}{https://github.com/google-research/vision\_transformer} and distributed under the \href{https://github.com/google-research/vision_transformer/blob/main/LICENSE}{Apache License 2.0}.

\item All the 124 datasets are publicly available via the Tensorflow Datasets image classification,  catalog.
Refer to \href{https://www.tensorflow.org/datasets/catalog/overview}{https://www.tensorflow.org/datasets/catalog/overview} for detailed information regarding each dataset licence and other metadata.
Table \ref{table:datasets} reports dataset splits used for the experiments and reference for each task.
\end{itemize}

We also publish the \method checkpoint resulting from the large-scale multitask experiment reported in Section~\ref{section:experiments}.
This checkpoint can be used for inference on any of the 124 learned image classification tasks, or for further analysis, or even to be extended with additional tasks or methods.
For information about the checkpoint and its license refer to: 
\href{https://github.com/google-research/google-research/tree/master/muNet}{https://github.com/google-research/google-research/tree/master/muNet}

The experiments reported in Sections \ref{section:preliminary} have been executed on a TPUv3 \citep{Jouppi2017IndatacenterPA} machine with 8 cores.
While, the large scale experiments reported in Section \ref{section:experiments} have been executed on a larger scale infrastructure using 32 TPUv4 chips in MegaCore mode,
by using the Pathways orchestration layer \citep{Barham2022PathwaysAD}, to accommodate for the increased memory requirements and scale.

\vspace{33pt}

\begin{table}[h]
    \centering
    \caption{Video representation of evolution of the large scale multitask system extended by each of the experiment segment described in Section~\ref{section:experiments}}
    \begin{tabular}{l|c}
         Segment & Video  \\
         \hline
         1: $\mu$2Net 2 task iterations on task-set A & \href{https://youtu.be/hDDCMpfCJ3s}{\small{\texttt{https://youtu.be/hDDCMpfCJ3s}}}  \\
         2: $\mu$2Net+ 2 task iterations on task-set A & \href{https://youtu.be/tf5tzuxbxx8}{\small{\texttt{https://youtu.be/tf5tzuxbxx8}}}  \\
         3: $\mu$2Net+ 1 task iteration to add task-set B & \href{https://youtu.be/jGkzXE2WLV0}{\small{\texttt{https://youtu.be/jGkzXE2WLV0}}}  \\
         4: $\mu$2Net+ 2 task iterations on task-set A+B & \href{https://youtu.be/CW_BAwkQ9e8}{\small{\texttt{https://youtu.be/CW\_BAwkQ9e8}}}  \\
    \end{tabular}
    \label{tab:videos}
\end{table}

\begin{figure}[h]
\centering
\includegraphics[width=1.\linewidth]{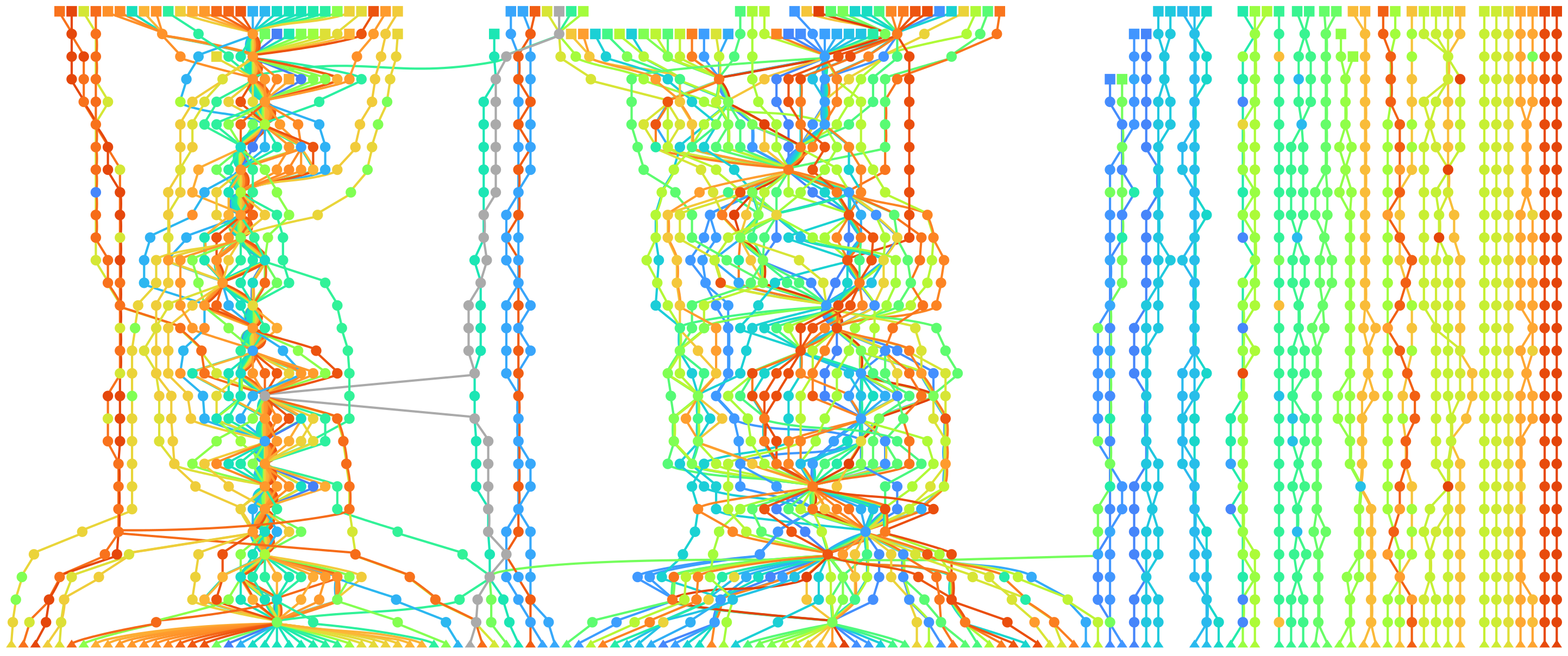}
\caption{Graph representing the architecture of the multitask system solving jointly 124 image classification tasks generated by the large scale continual learning experiments described in Section~\ref{section:experiments}.
Each task is identified with a unique color.
Bottom triangular nodes represent the data input of each task.
Top rectangular nodes represent the head layer of each task.
Each edges sequence of the same color connecting a task input to its head defines the layers sequence composing the model for each task.
Each path traverses a sequence of round nodes representing ViT L/16 internal layers in the following order from bottom to top: patch embedding, class token, position embedding and a variable number of transformer layers.
Internal nodes are represented with the color of the task on which the parameters of the corresponding layer were trained last.
Except for the gray nodes that have not received gradient updates from any of the 124 tasks and still carry the parameters of the root model that was loaded from a checkpoint of the ViT Large model pretrained on the imagenet-21k dataset as described by \cite{Gesmundo2022munet2}. See Table~\ref{tab:videos} for animations of the evolutionary process.
\\
The structure displays the formation of two main clusters of models/paths having a high degree of layers sharing.
These are models selected for tasks that can achieve peak quality with a lower degree of knowledge specialization.
Few models form smaller and disconnected clusters of paths.
Notice that knowledge sharing is effective even for models/paths that result in a disconnected sub-graph, since ancestors models may have been trained on multiple tasks whose current best model has fully branched in a separate disconnected structure during the evolutionary process.
Among these small clusters it is visible a high degree of sharing among related tasks.
For example, the leftmost disconnected subgraph of 3 paths  contains only characters classification tasks: \emph{bangla}, \emph{devanagari} and \emph{binary\_alpha\_digits} (see Table~\ref{table:datasets} for datasets details).
The majority of the disconnected subgraphs of 2 paths contain each a VTAB-Full task paired with the matching VTAB-1k "few-shots learning" version of the same task, from left to right: 1) \emph{clevr/count\_all} and \emph{clevr/count\_all}$_{1k}$,
2) \emph{dmlab} and \emph{dmlab}$_{1k}$,
3) \emph{dsprites/label\_orientation} and \emph{dsprites/label\_orientation}$_{1k}$,
4) \emph{eurosat} and \emph{eurosat}$_{1k}$,
5) \emph{resisc45} and \emph{resisc45}$_{1k}$,
6) \emph{smallnorb/label\_azimuth} and \emph{smallnorb/label\_azimuth}$_{1k}$.
}
\label{fig:deca-30}
\end{figure}

\begin{figure}[h]
\centering
\includegraphics[width=1.\linewidth]{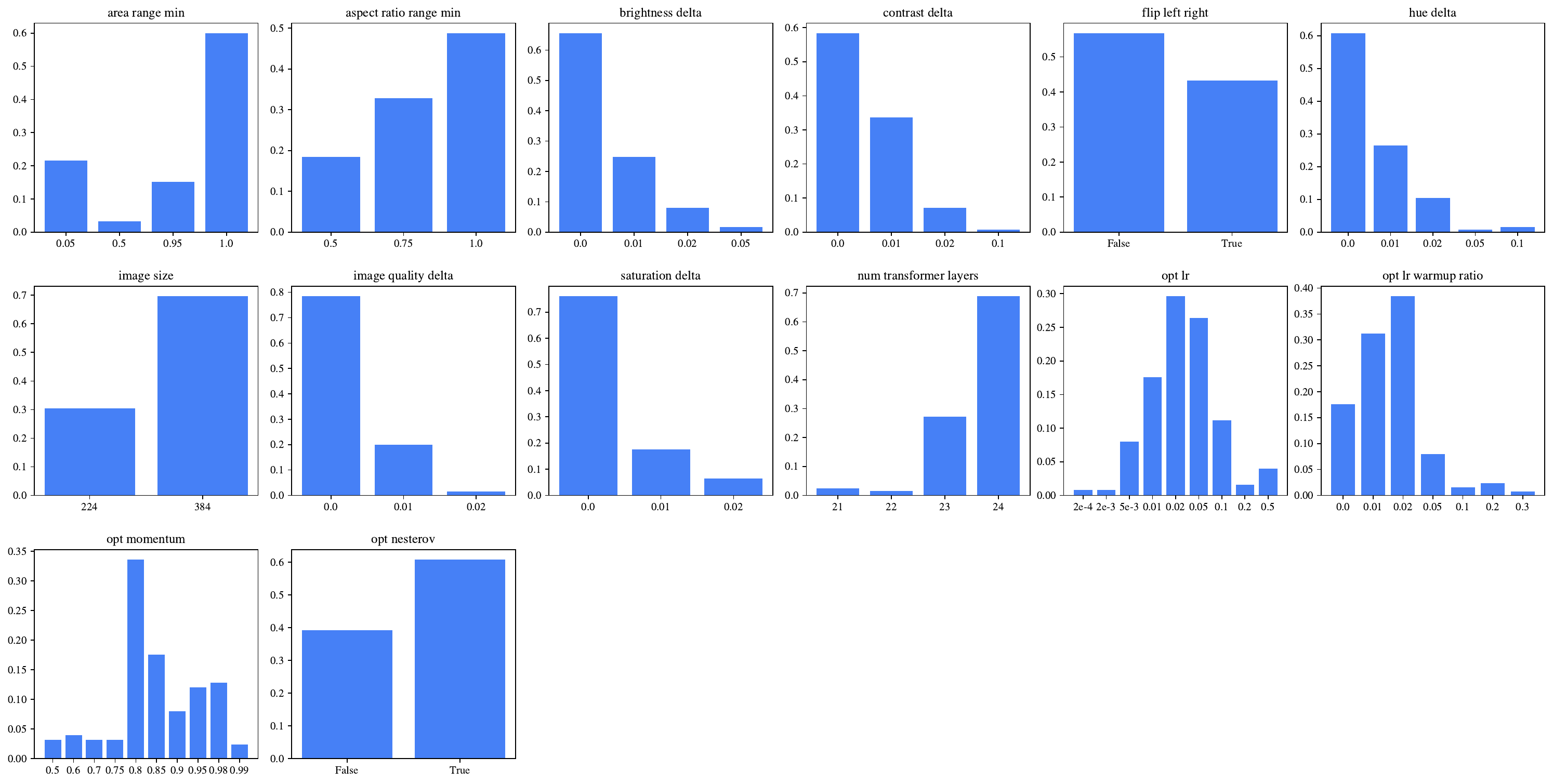}
\caption{Distributions of the hyperparameter values used by the 124 models included in the multitask system at the end of the sequence of experiments described in Section~\ref{section:experiments}.
}
\label{fig:dists}
\end{figure}

\begin{figure}[h]
\centering
\includegraphics[width=0.55\linewidth]{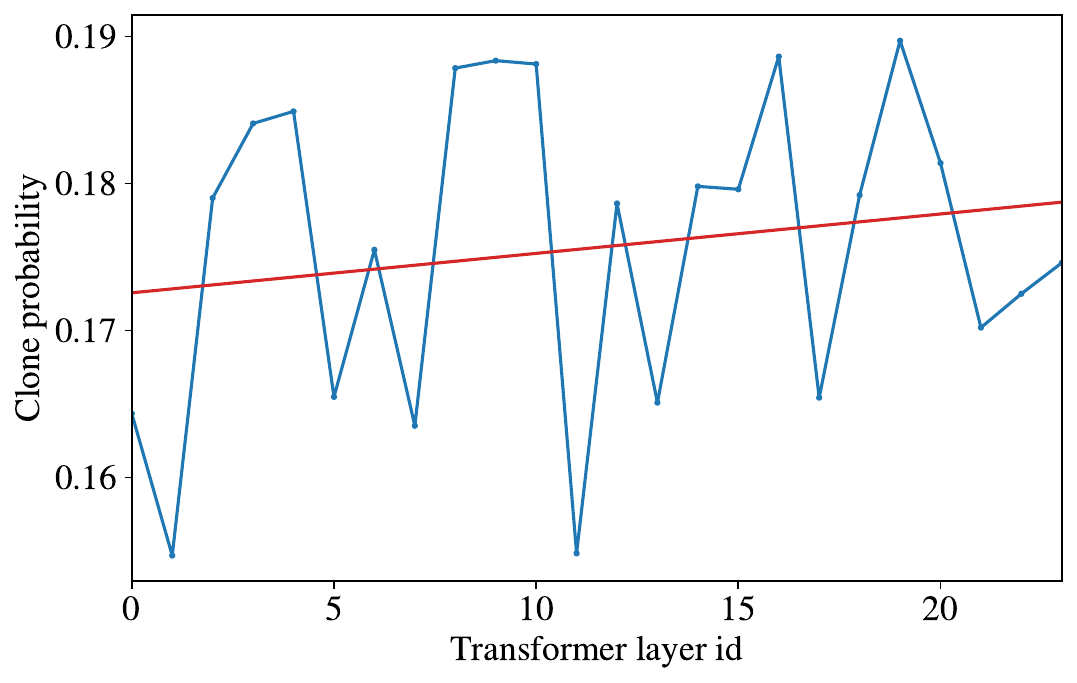}
\caption{Displays the mutation probabilities learned by the $\mu$ function for cloning transformer layers at different depth, lower layer ids correspond to transformer layers closer to the model input.
The displayed mutation probabilities are averaged across the values learned by the $\mu$ functions learned for the 124 models included in the multitask system at the end of the sequence of experiments described in Section~\ref{section:experiments}.
The red line is fitted to minimize the squared distance to the curve.
}
\label{fig:clone-prob}
\end{figure}

\begin{figure}[h]
\centering
\includegraphics[width=1.\linewidth]{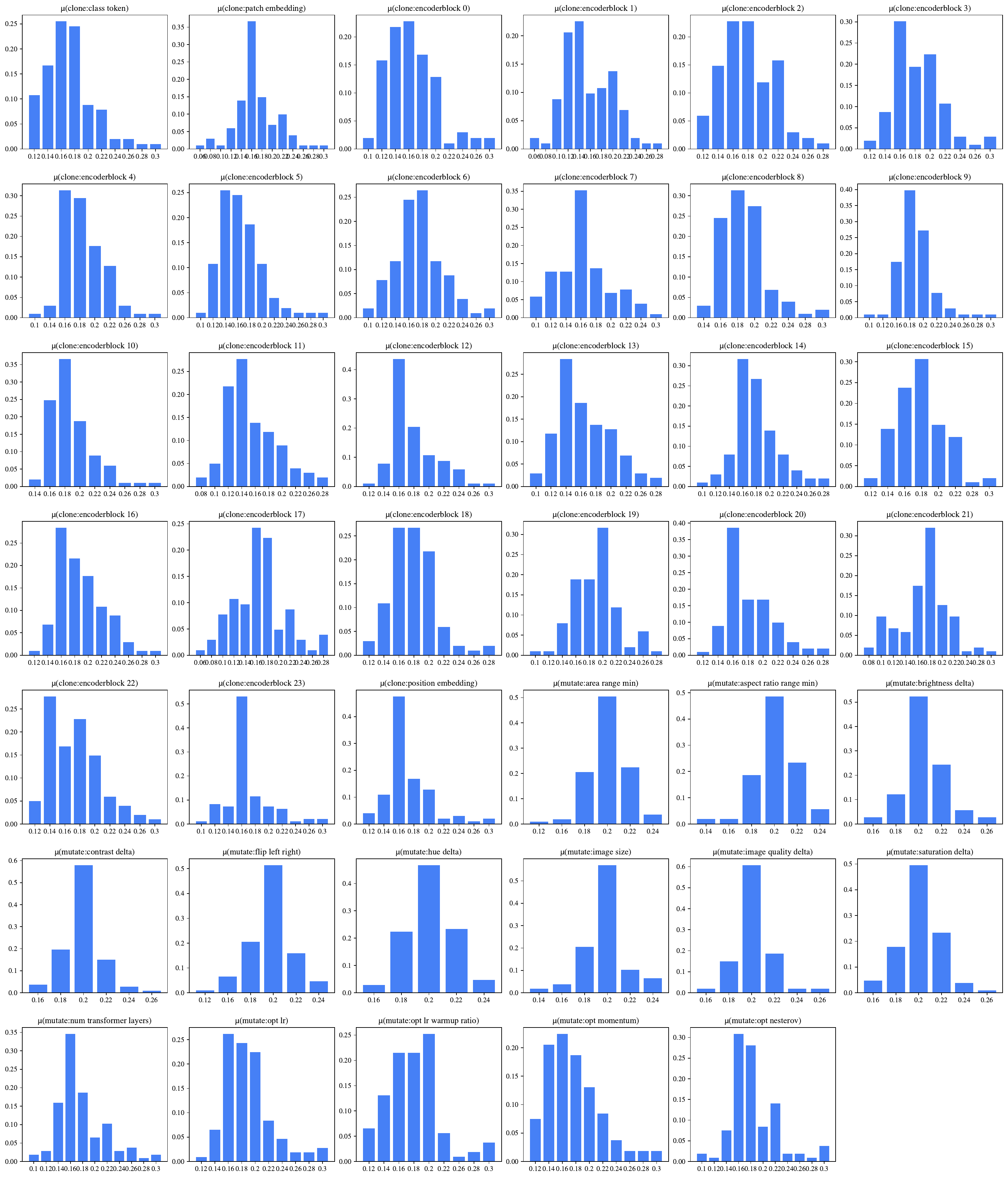}
\caption{Distributions of the $\mu(\cdot)$ values conditioned on different mutations of hyperparameters and clonable layers.
The histograms aggregates the values of the 124 models included in the multitask system at the end of the sequence of experiments described in Section~\ref{section:experiments}. 
}
\label{fig:dists_mu}
\end{figure}

\begin{table*}[h]
\small
\caption{Datasets details (part 1 of 3).
For each dataset used in the experiments, this table reports:
1) accuracy achieved on the test and validations sets by the large scale multitask model generated by the experiments described in Section~\ref{section:experiments}
2) dataset name indicative of the Tensorflow Datasets Catalogs identification string and linking to the corresponding catalog page,
3) train, validation and test data splits, represented with the \href{https://www.tensorflow.org/datasets/splits}{standard Tensorflow Datasets format} ("validation" has been abbreviated as "val").
4) corresponding scientific publication reference.
Datasets are listed in the order of introduction into the 
system.
\\\hspace{\textwidth}
Notes:
\\\hspace{\textwidth}
[1] The test split of the \href{https://www.tensorflow.org/datasets/catalog/imagenet_v2}{imagenet\_v2} dataset is used as validation set for \href{https://www.tensorflow.org/datasets/catalog/imagenet2012}{imagenet2012}. 
\\\hspace{\textwidth}
[2] The test split of the \href{https://www.tensorflow.org/datasets/catalog/cifar10_1}{cifar10\_1} dataset is used as validation set for \href{https://www.tensorflow.org/datasets/catalog/cifar10}{cifar10}.
\\\hspace{\textwidth}
[3] The VTAB-full benchmark also includes the cifar100 task. Cifar100 has been introduced to the system as part of the initial benchmark. In the VTAB-full results tables we refer to the top 1 test accuracy achieved in the latest cifar100 training iteration without retraining it as part of the VTAB-full active training iteration.
\\\hspace{\textwidth}
[4] The definition for the VTAB standard and additional tasks has been sourced from \href{https://github.com/google-research/task_adaptation/tree/master/task_adaptation/data}{https://github.com/google-research/task\_adaptation/tree/master/task\_adaptation/data}.
\\\hspace{\textwidth}
[5] VTAB additional task, not included in the standard scoring set. These tasks were added to further scale the system and analyze transfer across related tasks.
}
\label{table:datasets}
\centering
\setlength\tabcolsep{1pt}
\hspace*{-13.32pt}
\begin{tabular}{lcccccc}
    \toprule
    &\multicolumn{2}{c}{Accuracy (\%)}
    & \multicolumn{3}{c}{Splits}                   \\
    \cmidrule(r){2-3}
    \cmidrule(r){4-6}
    Name & Val. & Test   & Train & Val. & Test  & Reference\\
    \midrule
    \midrule

\multicolumn{7}{c}{\textbf{Task-set A}} \\
\midrule

\href{https://www.tensorflow.org/datasets/catalog/imagenet2012}{imagenet2012}
&  78.54  &  86.66
& train & {\tiny \href{https://www.tensorflow.org/datasets/catalog/imagenet_v2}{imagenet\_v2}:}test$^{[1]}$ & val
& \citep{Russakovsky2015ImageNetLS}
\\
\href{https://www.tensorflow.org/datasets/catalog/cifar100}{cifar100}
&  96.88  &  94.94
& train[{\tiny:98\%}] & train[{\tiny98\%:}] & test
& \citep{Krizhevsky2009LearningML}
\\
\href{https://www.tensorflow.org/datasets/catalog/cifar10}{cifar10}
&  98.81  &  99.48
& train & {\tiny \href{https://www.tensorflow.org/datasets/catalog/cifar10_1}{cifar10\_1}:}test$^{[2]}$ & test
& \citep{Krizhevsky2009LearningML}
\\

\midrule
\multicolumn{7}{c}{VTAB-full benchmark$^{[3][4]}$} \\

\href{https://www.tensorflow.org/datasets/catalog/caltech101}{caltech101}
&  98.67  &  95.94
& train[{\tiny:2754}] & train[{\tiny2754:}] & test
& \citep{FeiFei2004LearningGV}
\\
\href{https://www.tensorflow.org/datasets/catalog/dtd}{dtd}
&  82.63  &  82.23
& train & val & test
& \citep{Cimpoi2014DescribingTI}
\\
\href{https://www.tensorflow.org/datasets/catalog/oxford_flowers102}{oxford\_flowers102}
&  99.79  &  99.48
& train & val & test
& \hspace{-10pt}\citep{Nilsback2008AutomatedFC}
\\
\href{https://www.tensorflow.org/datasets/catalog/oxford_iiit_pet}{oxford\_iiit\_pet}
&  98.09  &  95.50
& train[{\tiny:2944}] & train[{\tiny2944:}] & test
& \citep{Parkhi2012CatsAD}
\\
\href{https://www.tensorflow.org/datasets/catalog/sun397}{sun397}
&  84.71  &  84.32
& train & val & test
& \citep{Xiao2010SUNDL}
\\
\href{https://www.tensorflow.org/datasets/catalog/svhn_cropped}{svhn\_cropped}
&  97.47  &  97.50
& train[{\tiny:65931}] & train[{\tiny65931:}] & test
& \citep{Netzer2011ReadingDI}
\\
\href{https://www.tensorflow.org/datasets/catalog/patch_camelyon}{patch\_camelyon}
&  92.82  &  91.14
& train & val & test
& \citep{Veeling2018RotationEC}
\\
\href{https://www.tensorflow.org/datasets/catalog/eurosat#eurosatrgb_default_config}{eurosat/rgb}
&  99.27  &  99.22
& train[{\tiny:16200}] & train[{\tiny16200:21600}] & train[{\tiny21600:}]
& \citep{Helber2019EuroSATAN}
\\
\href{https://www.tensorflow.org/datasets/catalog/resisc45}{resisc45}
&  97.80  &  97.21
& train[{\tiny:18900}] & train[{\tiny18900:25200}] & train[{\tiny25200:}]
& \citep{Cheng2017RemoteSI}
\\
\multicolumn{2}{l}{
diabetic\_retinopathy\_detection/...
}
\\
\ \ \ \ \href{https://www.tensorflow.org/datasets/catalog/diabetic_retinopathy_detection/#diabetic_retinopathy_detectionbtgraham-300}{btgraham-300}
&  85.22  &  83.75
& train & val & test
& \hspace{-10pt}\citep{kaggle-diabetic-retinopathy}
\\
\href{https://www.tensorflow.org/datasets/catalog/clevr}{clevr/count\_cylinders}$^{[5]}$
&  99.65  &  99.47
& train[{\tiny:63000}] & train[{\tiny63000:}] & val
& \citep{Johnson2017CLEVRAD}
\\
\href{https://www.tensorflow.org/datasets/catalog/clevr}{clevr/count\_all}
&  99.97  &  99.90
& train[{\tiny:63000}] & train[{\tiny63000:}] & val
& \citep{Johnson2017CLEVRAD}
\\
\href{https://www.tensorflow.org/datasets/catalog/clevr}{clevr/closest\_object\_distance}
&  94.63  &  94.05
& train[{\tiny:63000}] & train[{\tiny63000:}] & val
& \citep{Johnson2017CLEVRAD}
\\
\href{https://www.tensorflow.org/datasets/catalog/dmlab}{dmlab}
&  77.01  &  76.92
& train & val & test
& \citep{Zhai2019TheVT}
\\
\href{https://www.tensorflow.org/datasets/catalog/dsprites}{dsprites/label\_x\_position}
&  99.99  &  99.98
& train[{\tiny:589824}] & train[{\tiny589824:663552}] & train[{\tiny663552:}]
& \citep{Klindt2021TowardsND}
\\
\href{https://www.tensorflow.org/datasets/catalog/dsprites}{dsprites/label\_orientation}
&  96.78  &  96.44
& train[{\tiny:589824}] & train[{\tiny589824:663552}] & train[{\tiny663552:}]
& \citep{Klindt2021TowardsND}
\\
\href{https://www.tensorflow.org/datasets/catalog/kitti}{kitti/closest\_object\_distance}$^{[5]}$
&  83.57  &  78.48
& train & val & test
& \citep{Geiger2012AreWR}
\\
\href{https://www.tensorflow.org/datasets/catalog/kitti}{kitti/count\_vehicles}$^{[5]}$
&  92.14  &  76.93
& train & val & test
& \citep{Geiger2012AreWR}
\\
\href{https://www.tensorflow.org/datasets/catalog/kitti}{kitti/closest\_vehicle\_distance}
&  89.76  &  82.28
& train & val & test
& \citep{Geiger2012AreWR}
\\
\href{https://www.tensorflow.org/datasets/catalog/smallnorb}{smallnorb/label\_category}$^{[5]}$
&  99.45  &  99.38
& train & test[{\tiny:50\%}] & test[{\tiny50\%:}]
& \citep{LeCun2004LearningMF}
\\
\href{https://www.tensorflow.org/datasets/catalog/smallnorb}{smallnorb/label\_lighting}$^{[5]}$
&  99.93  &  99.89
& train & test[{\tiny:50\%}] & test[{\tiny50\%:}]
& \citep{LeCun2004LearningMF}
\\
\href{https://www.tensorflow.org/datasets/catalog/smallnorb}{smallnorb/label\_azimuth}
&  35.40  &  34.44
& train & test[{\tiny:50\%}] & test[{\tiny50\%:}]
& \citep{LeCun2004LearningMF}
\\
\href{https://www.tensorflow.org/datasets/catalog/smallnorb}{smallnorb/label\_elevation}
&  96.22  &  96.26
& train & test[{\tiny:50\%}] & test[{\tiny50\%:}]
& \citep{LeCun2004LearningMF}
\\

\midrule

\multicolumn{7}{c}{Continues in Table~\ref{table:datasets2} \dots}  \\
\bottomrule
  \end{tabular}
\end{table*}

\begin{table*}[h]
\small
\caption{Datasets details (part 2 of 3).}
\label{table:datasets2}
\centering
\setlength\tabcolsep{1pt}
\hspace*{-9.82pt}
\begin{tabular}{lcccccc}
    \toprule
    & \multicolumn{2}{c}{Accuracy (\%)}
    & \multicolumn{3}{c}{Splits}                   \\
    \cmidrule(r){2-3}
    \cmidrule(r){4-6}
    Name & Val. & Test & Train & Val. & Test  & Reference\\
    \midrule
\multicolumn{7}{c}{\dots Continues from Table~\ref{table:datasets} }\\
\midrule

\multicolumn{7}{c}{Visual domain decathlon benchmark} \\

\multicolumn{2}{l}{
visual\_domain\_decathlon/...
}
&
\\
\ \ \ \ \href{https://www.tensorflow.org/datasets/catalog/visual_domain_decathlon#visual_domain_decathlonimagenet12}{imagenet12}
&  89.47  &  89.69
& train & val[{\tiny:50\%}] & val[{\tiny50\%:}]
& \citep{hakanbilensylvestrerebuffitomasjakab2017}
\\
\ \ \ \ \href{https://www.tensorflow.org/datasets/catalog/visual_domain_decathlon#visual_domain_decathlonsvhn}{svhn}
&  98.75  &  98.57
& train & val[{\tiny:50\%}] & val[{\tiny50\%:}]
& \citep{hakanbilensylvestrerebuffitomasjakab2017}
\\
\ \ \ \ \href{https://www.tensorflow.org/datasets/catalog/visual_domain_decathlon#visual_domain_decathloncifar100}{cifar100}
&  97.79  &  98.00
& train & val[{\tiny:50\%}] & val[{\tiny50\%:}]
& \citep{hakanbilensylvestrerebuffitomasjakab2017}
\\
\ \ \ \ \href{https://www.tensorflow.org/datasets/catalog/visual_domain_decathlon#visual_domain_decathlongtsrb}{gtsrb}
&  100  &  99.95
& train & val[{\tiny:50\%}] & val[{\tiny50\%:}]
& \citep{hakanbilensylvestrerebuffitomasjakab2017}
\\
\ \ \ \ \href{https://www.tensorflow.org/datasets/catalog/visual_domain_decathlon#visual_domain_decathlondaimlerpedcls}{daimlerpedcls}
&  100  &  100
& train & val[{\tiny:50\%}] & val[{\tiny50\%:}]
& \citep{hakanbilensylvestrerebuffitomasjakab2017}
\\
\ \ \ \ \href{https://www.tensorflow.org/datasets/catalog/visual_domain_decathlon#visual_domain_decathlonomniglot}{omniglot}
&  88.70  &  87.99
& train & val[{\tiny:50\%}] & val[{\tiny50\%:}]
& \citep{hakanbilensylvestrerebuffitomasjakab2017}
\\
\ \ \ \ \href{https://www.tensorflow.org/datasets/catalog/visual_domain_decathlon#visual_domain_decathlonucf101}{ucf101}
&  85.96  &  87.70
& train & val[{\tiny:50\%}] & val[{\tiny50\%:}]
& \citep{hakanbilensylvestrerebuffitomasjakab2017}
\\
\ \ \ \ \href{https://www.tensorflow.org/datasets/catalog/visual_domain_decathlon#visual_domain_decathlonaircraft_default_config}{aircraft}
&  72.95  &  70.01
& train & val[{\tiny:50\%}] & val[{\tiny50\%:}]
& \citep{hakanbilensylvestrerebuffitomasjakab2017}
\\
\ \ \ \ \href{https://www.tensorflow.org/datasets/catalog/visual_domain_decathlon#visual_domain_decathlondtd}{dtd}
&  73.11  &  75.00
& train & val[{\tiny:50\%}] & val[{\tiny50\%:}]
& \citep{hakanbilensylvestrerebuffitomasjakab2017}
\\
\ \ \ \ \href{https://www.tensorflow.org/datasets/catalog/visual_domain_decathlon#visual_domain_decathlonvgg-flowers}{vgg-flowers}
&  99.37  &  99.41
& train & val[{\tiny:50\%}] & val[{\tiny50\%:}]
& \citep{hakanbilensylvestrerebuffitomasjakab2017}
\\
\midrule

\multicolumn{7}{c}{Multitask Character Classification Benchmark} \\

\href{https://www.tensorflow.org/datasets/catalog/emnist#emnistdigits}{emnist/digits}
&  99.86  &  99.81
& train[{\tiny5\%:}] & train[{\tiny:5\%}] & test
& \citep{Cohen2017EMNISTEM}
\\
\href{https://www.tensorflow.org/datasets/catalog/emnist#emnistletters}{emnist/letters}
&  96.57  &  95.03
&train[{\tiny5\%:}] & train[{\tiny:5\%}] & test
& \citep{Cohen2017EMNISTEM}
\\
\href{https://www.tensorflow.org/datasets/catalog/kmnist}{kmnist}
&  99.79  &  98.44
& train[{\tiny5\%:}] & train[{\tiny:5\%}] & test
& \citep{Clanuwat2018DeepLF}
\\
\href{https://www.tensorflow.org/datasets/catalog/mnist}{mnist}
&  99.83  &  99.72
& train[{\tiny5\%:}] & train[{\tiny:5\%}] & test
& \citep{LeCun1998GradientbasedLA}
\\
\href{https://www.tensorflow.org/datasets/catalog/omniglot}{omniglot}
&  99.89  &  99.90
& train & small1 & small2
& \citep{Lake2015HumanlevelCL}
\\
\href{https://www.tensorflow.org/datasets/catalog/cmaterdb#cmaterdbbangla_default_config}{cmaterdb/bangla}
&  99.89  &  99.00
& train[{\tiny20\%:}] & train[{\tiny:20\%}] & test
& \citep{Das2012AGA,Das2012ASF}
\\
\href{https://www.tensorflow.org/datasets/catalog/cmaterdb#cmaterdbdevanagari}{cmaterdb/devanagari}
&  100  &  98.20
& train[{\tiny20\%:}] & train[{\tiny:20\%}] & test
& \citep{Das2012AGA,Das2012ASF}
\\
\href{https://www.tensorflow.org/datasets/catalog/cmaterdb#cmaterdbtelugu}{cmaterdb/telugu}
&  100  &  99.40
& train[{\tiny20\%:}] & train[{\tiny:20\%}] & test
& \citep{Das2012AGA,Das2012ASF}
\\

\midrule

\multicolumn{7}{c}{VTAB 1k benchmark$^{[4]}$} \\

\href{https://www.tensorflow.org/datasets/catalog/caltech101}{caltech101}
&  98.97  &  89.88
& train[{\tiny:800}] & train[{\tiny2754:2954}] & test
& \citep{FeiFei2004LearningGV}
\\
\href{https://www.tensorflow.org/datasets/catalog/cifar100}{cifar100}
&  96.92  &  92.85
& train[{\tiny:800}] & train[{\tiny45000:45200}] & test
& \citep{Krizhevsky2009LearningML}
\\
\href{https://www.tensorflow.org/datasets/catalog/cifar10}{cifar10}
&  100  &  99.23
& train[{\tiny:800}] & train[{\tiny45000:45200}] & test
& \citep{Krizhevsky2009LearningML}
\\
\href{https://www.tensorflow.org/datasets/catalog/dtd}{dtd}
&  82.05  &  77.87
& train[{\tiny:800}] & val[{\tiny:200}] & test
& \citep{Cimpoi2014DescribingTI}
\\
\href{https://www.tensorflow.org/datasets/catalog/oxford_flowers102}{oxford\_flowers102}
&  100  &  99.33
& train[{\tiny:800}] & val[{\tiny:200}] & test
& \hspace{-10pt}\citep{Nilsback2008AutomatedFC}
\\
\href{https://www.tensorflow.org/datasets/catalog/oxford_iiit_pet}{oxford\_iiit\_pet}
&  97.44  &  93.51
& train[{\tiny:800}] & train[{\tiny2944:3144}] & test
& \citep{Parkhi2012CatsAD}
\\
\href{https://www.tensorflow.org/datasets/catalog/sun397}{sun397}
&  66.15  &  60.93
& train[{\tiny:800}] & val[{\tiny:200}] & test
& \citep{Xiao2010SUNDL}
\\
\href{https://www.tensorflow.org/datasets/catalog/svhn_cropped}{svhn\_cropped}
&  98.00  &  97.47
& train[{\tiny:800}] & train[{\tiny65931:66131}] & test
& \citep{Netzer2011ReadingDI}
\\
\href{https://www.tensorflow.org/datasets/catalog/patch_camelyon}{patch\_camelyon}
&  96.41  &  91.55
& train[{\tiny:800}] & val[{\tiny:200}] & test
& \citep{Veeling2018RotationEC}
\\
\href{https://www.tensorflow.org/datasets/catalog/eurosat#eurosatrgb_default_config}{eurosat/rgb}
 &  99.49  &  98.56
& train[{\tiny:800}] & train[{\tiny16200:16400}] & train[{\tiny21600:}]
& \citep{Helber2019EuroSATAN}
\\
\href{https://www.tensorflow.org/datasets/catalog/resisc45}{resisc45}
&  97.50  &  95.33
& train[{\tiny:800}] & train[{\tiny18900:19100}] & train[{\tiny25200:}]
& \citep{Cheng2017RemoteSI}
\\
\multicolumn{2}{l}{
diabetic\_retinopathy\_detection/...
}
\\
\ \ \ \ \href{https://www.tensorflow.org/datasets/catalog/diabetic_retinopathy_detection/#diabetic_retinopathy_detectionbtgraham-300}{btgraham-300}
&  88.50  &  82.77
& train[{\tiny:800}] & val[{\tiny:200}] & test
& \hspace{-10pt}\citep{kaggle-diabetic-retinopathy}
\\
\href{https://www.tensorflow.org/datasets/catalog/clevr}{clevr/count\_cylinders}$^{[5]}$
 &  99.49  &  99.01
& train[{\tiny:800}] & train[{\tiny63000:63200}] & val
& \citep{Johnson2017CLEVRAD}
\\
\href{https://www.tensorflow.org/datasets/catalog/clevr}{clevr/count\_all}
&  100  &  99.88
& train[{\tiny:800}] & train[{\tiny63000:63200}] & val
& \citep{Johnson2017CLEVRAD}
\\
\href{https://www.tensorflow.org/datasets/catalog/clevr}{clevr/closest\_object\_distance}
&  92.00  &  90.64
& train[{\tiny:800}] & train[{\tiny63000:63200}] & val
& \citep{Johnson2017CLEVRAD}
\\
\href{https://www.tensorflow.org/datasets/catalog/dmlab}{dmlab}
&  77.44  &  74.71
& train[{\tiny:800}] & val[{\tiny:200}] & test
& \citep{Zhai2019TheVT}
\\
\href{https://www.tensorflow.org/datasets/catalog/dsprites}{dsprites/label\_x\_position}
&  100  &  99.43
& train[{\tiny:800}] & train[{\tiny589824:590024}] & train[{\tiny663552:}]
& \citep{Klindt2021TowardsND}
\\
\href{https://www.tensorflow.org/datasets/catalog/dsprites}{dsprites/label\_orientation}
&  97.50  &  96.30
& train[{\tiny:800}] & train[{\tiny589824:590024}] & train[{\tiny663552:}]
& \citep{Klindt2021TowardsND}
\\
\href{https://www.tensorflow.org/datasets/catalog/kitti}{kitti/closest\_object\_distance}$^{[5]}$
&  84.50  &  78.34
& train[{\tiny:800}] & val[{\tiny:200}] & test
& \citep{Geiger2012AreWR}
\\
\href{https://www.tensorflow.org/datasets/catalog/kitti}{kitti/count\_vehicles}$^{[5]}$
&  93.85  &  69.34
& train[{\tiny:800}] & val[{\tiny:200}] & test
& \citep{Geiger2012AreWR}
\\
\href{https://www.tensorflow.org/datasets/catalog/kitti}{kitti/closest\_vehicle\_distance}
&  88.00  &  82.14
& train[{\tiny:800}] & val[{\tiny:200}] & test
& \citep{Geiger2012AreWR}
\\
\href{https://www.tensorflow.org/datasets/catalog/smallnorb}{smallnorb/label\_category}$^{[5]}$
&  100  &  97.67
& train[{\tiny:800}] & test[{\tiny:200}] & test[{\tiny50\%:}]
& \citep{LeCun2004LearningMF}
\\
\href{https://www.tensorflow.org/datasets/catalog/smallnorb}{smallnorb/label\_lighting}$^{[5]}$
 &  100  &  98.30
& train[{\tiny:800}] & test[{\tiny:200}] & test[{\tiny50\%:}]
& \citep{LeCun2004LearningMF}
\\
\href{https://www.tensorflow.org/datasets/catalog/smallnorb}{smallnorb/label\_azimuth}
&  37.00  &  33.75
& train[{\tiny:800}] & test[{\tiny:200}] & test[{\tiny50\%:}]
& \citep{LeCun2004LearningMF}
\\
\href{https://www.tensorflow.org/datasets/catalog/smallnorb}{smallnorb/label\_elevation}
&  92.50  &  92.83
& train[{\tiny:800}] & test[{\tiny:200}] & test[{\tiny50\%:}]
& \citep{LeCun2004LearningMF}
\\
\midrule
\multicolumn{7}{c}{Continues in Table~\ref{table:datasets3} \dots}  \\

    \bottomrule
  \end{tabular}
\end{table*}

\clearpage

\begin{table*}[h]
\small
\caption{Datasets details (part 3 of 3).}
\label{table:datasets3}
\centering
\setlength\tabcolsep{1pt}
\begin{tabular}{lcccccc}
    \toprule
    & \multicolumn{2}{c}{Accuracy (\%)}
    & \multicolumn{3}{c}{Splits}                   \\
    \cmidrule(r){2-3}
    \cmidrule(r){4-6}
    Name & Val. & Test & Train & Val. & Test  & Reference\\
    \midrule
\multicolumn{7}{c}{\dots Continues from Table~\ref{table:datasets2} }\\
\midrule
\midrule
\multicolumn{7}{c}{\textbf{Task-set B}}\\
\midrule

\href{https://www.tensorflow.org/datasets/catalog/beans}{beans}
& 100 &  94.53 
 & train & val & test & 
 \citep{beansdata}
 \\
\href{https://www.tensorflow.org/datasets/catalog/binary_alpha_digits}{binary\_alpha\_digits}
&  91.43  &  85.71
 & train[{\tiny10\%:}] & train[{\tiny5\%:10\%}] & train[{\tiny:5\%}] &
 $-$ 
 \\
\href{https://www.tensorflow.org/datasets/catalog/caltech_birds2010}{caltech\_birds2010}
&  98.00  &  89.48
 & train[{\tiny5\%:}] & train[{\tiny:5\%}] & test & 
 \citep{Welinder2010CaltechUCSDB2}
 \\
\href{https://www.tensorflow.org/datasets/catalog/caltech_birds2011}{caltech\_birds2011}
&  93.00  &  90.94
 & train[{\tiny5\%:}] & train[{\tiny:5\%}] & test & 
 \citep{Welinder2010CaltechUCSDB2}
 \\
\href{https://www.tensorflow.org/datasets/catalog/cars196}{cars196}
&  89.23  &  87.18
 & train[{\tiny5\%:}] & train[{\tiny:5\%}] & test & 
 \citep{Krause20133DOR}
 \\
\href{https://www.tensorflow.org/datasets/catalog/cassava}{cassava}
&  92.31  &  91.78
 & train & val & test & 
 \citep{Mwebaze2019iCassava2V}
 \\
\href{https://www.tensorflow.org/datasets/catalog/cats_vs_dogs}{cats\_vs\_dogs}
&  100 &  99.83
 & train[{\tiny10\%:}] & train[{\tiny5\%:10\%}] & train[{\tiny:5\%}] &
 \citep{Elson2007AsirraAC}
 \\
\href{https://www.tensorflow.org/datasets/catalog/citrus_leaves}{citrus\_leaves}
&  100  &  93.33
 & train[{\tiny10\%:}] & train[{\tiny5\%:10\%}] & train[{\tiny:5\%}] &
 \citep{Rauf2019ACF}
 \\
\href{https://www.tensorflow.org/datasets/catalog/colorectal_histology}{colorectal\_histology}
&  99.17  &  98.40
 & train[{\tiny10\%:}] & train[{\tiny5\%:10\%}] & train[{\tiny:5\%}] &
 \citep{Kather2016MulticlassTA}
 \\
\multicolumn{2}{l}{
controlled\_noisy\_web\_labels/...
}
 &
 \\
\ \ \ \ \href{https://www.tensorflow.org/datasets/catalog/controlled_noisy_web_labels#mini_imagenet_red}{mini\_imagenet\_red}
&  96.35  &  95.04
 & train\_00 & val[{\tiny:50\%}] & val[{\tiny50\%:}] & 
 \citep{Jiang2020BeyondSN}
 \\
\ \ \ \ \href{https://www.tensorflow.org/datasets/catalog/controlled_noisy_web_labels#mini_imagenet_blue}{mini\_imagenet\_blue}
&  96.35  &  95.24
 & train\_00 & val[{\tiny:50\%}] & val[{\tiny50\%:}] &  \citep{Jiang2020BeyondSN}
 \\
\multicolumn{2}{l}{
curated\_breast\_imaging\_ddsm/...
}
 &
 \\
\ \ \ \ \href{https://www.tensorflow.org/datasets/catalog/curated_breast_imaging_ddsm#patches}{patches}
&  70.66  &  67.49
 & train & val & test & 
 \citep{Clark2013TheCI}
 \\
cycle\_gan/...
 &
 \\
\ \ \ \ \href{https://www.tensorflow.org/datasets/catalog/cycle_gan#apple2orange}{apple2orange}
 &  100  &  98.83
 &  {\tiny trainA+B[10\%:]} & {\tiny trainA+B[:10\%]} & {\tiny testA+B}&  
 \citep{Zhu2017UnpairedIT}
 \\
\ \ \ \ \href{https://www.tensorflow.org/datasets/catalog/cycle_gan#summer2winter_yosemite}{summer2winter} 
 & 95.71  &  90.49
 &  {\tiny trainA+B[10\%:]} & {\tiny trainA+B[:10\%]} & {\tiny testA+B}&
 \citep{Zhu2017UnpairedIT}
 \\
\ \ \ \ \href{https://www.tensorflow.org/datasets/catalog/cycle_gan#horse2zebra}{horse2zebra}
&  100  &  99.23
 &  {\tiny trainA+B[10\%:]} & {\tiny trainA+B[:10\%]} & {\tiny testA+B}&
 \citep{Zhu2017UnpairedIT}
 \\
\ \ \ \ \href{https://www.tensorflow.org/datasets/catalog/cycle_gan#monet2photo}{monet2photo}
&  100  &  100
 &  {\tiny trainA+B[10\%:]} & {\tiny trainA+B[:10\%]} & {\tiny testA+B}&
 \citep{Zhu2017UnpairedIT}
 \\
\ \ \ \ \href{https://www.tensorflow.org/datasets/catalog/cycle_gan#cezanne2photo}{cezanne2photo}
&  100  &  100
 &  {\tiny trainA+B[10\%:]} & {\tiny trainA+B[:10\%]} & {\tiny testA+B}&
 \citep{Zhu2017UnpairedIT}
 \\
\ \ \ \ \href{https://www.tensorflow.org/datasets/catalog/cycle_gan#ukiyoe2photo}{ukiyoe2photo}
&  100  &  99.70
 &  {\tiny trainA+B[10\%:]} & {\tiny trainA+B[:10\%]} & {\tiny testA+B}&
 \citep{Zhu2017UnpairedIT}
 \\
\ \ \ \ \href{https://www.tensorflow.org/datasets/catalog/cycle_gan#vangogh2photo}{vangogh2photo}
&  100  &  100
 &  {\tiny trainA+B[10\%:]} & {\tiny trainA+B[:10\%]} & {\tiny testA+B}&
 \citep{Zhu2017UnpairedIT}
 \\
\ \ \ \ \href{https://www.tensorflow.org/datasets/catalog/cycle_gan#maps}{maps}
&  100  &  100
 &  {\tiny trainA+B[10\%:]} & {\tiny trainA+B[:10\%]} & {\tiny testA+B}&
 \citep{Zhu2017UnpairedIT}
 \\
\ \ \ \ \href{https://www.tensorflow.org/datasets/catalog/cycle_gan#cityscapes}{cityscapes}
&  100  &  100
 &  {\tiny trainA+B[10\%:]} & {\tiny trainA+B[:10\%]} & {\tiny testA+B}&
 \citep{Zhu2017UnpairedIT}
 \\
\ \ \ \ \href{https://www.tensorflow.org/datasets/catalog/cycle_gan#facades}{facades}
&  100  &  100
 &  {\tiny trainA+B[10\%:]} & {\tiny trainA+B[:10\%]} & {\tiny testA+B}&
 \citep{Zhu2017UnpairedIT}
 \\
\ \ \ \ \href{https://www.tensorflow.org/datasets/catalog/cycle_gan#iphone2dslr_flower}{iphone2dslr\_flower}
&  98.99  &  94.66
 &  {\tiny trainA+B[10\%:]} & {\tiny trainA+B[:10\%]} & {\tiny testA+B}&
 \citep{Zhu2017UnpairedIT}
 \\
\href{https://www.tensorflow.org/datasets/catalog/deep_weeds}{deep\_weeds}
&  98.79  &  98.06
 & train[{\tiny10\%:}] & train[{\tiny5\%:10\%}] & train[{\tiny:5\%}] &
 \citep{Olsen2019DeepWeedsAM}
 \\
\href{https://www.tensorflow.org/datasets/catalog/domainnet#real}{domainnet/real}
&  90.90  &  90.25
 & train[{\tiny5\%:}] & train[{\tiny:5\%}] & test &
 \citep{Peng2019MomentMF}
 \\
\href{https://www.tensorflow.org/datasets/catalog/domainnet#painting}{domainnet/painting}
&  82.62  &  82.11
 & train[{\tiny5\%:}] & train[{\tiny:5\%}] & test &
 \citep{Peng2019MomentMF}
 \\
\href{https://www.tensorflow.org/datasets/catalog/domainnet#clipart}{domainnet/clipart}
&  87.16  &  85.35
 & train[{\tiny5\%:}] & train[{\tiny:5\%}] & test &
 \citep{Peng2019MomentMF}
 \\
\href{https://www.tensorflow.org/datasets/catalog/domainnet#quickdraw}{domainnet/quickdraw}
&  74.34  &  73.78
 & train[{\tiny5\%:}] & train[{\tiny:5\%}] & test & 
 \citep{Peng2019MomentMF}
 \\
\href{https://www.tensorflow.org/datasets/catalog/domainnet#infograph}{domainnet/infograph}
&  56.28  &  55.28
 & train[{\tiny5\%:}] & train[{\tiny:5\%}] & test &
 \citep{Peng2019MomentMF}
 \\
\href{https://www.tensorflow.org/datasets/catalog/domainnet#sketch}{domainnet/sketch}
&  78.70  &  78.51
 & train[{\tiny5\%:}] & train[{\tiny:5\%}] & test &
 \citep{Peng2019MomentMF}
 \\
\href{https://www.tensorflow.org/datasets/catalog/food101}{food101}
&  94.58  &  91.47
 & train[{\tiny5\%:}] & val & train[{\tiny:5\%}] &
 \citep{Bossard2014Food101M}
 \\
\href{https://www.tensorflow.org/datasets/catalog/horses_or_humans}{horses\_or\_humans}
&  100  &  99.61
 & train[{\tiny5\%:}] & train[{\tiny:5\%}] & test &
 \citep{horses_or_humans}
 \\
\href{https://www.tensorflow.org/datasets/catalog/i_naturalist2017}{i\_naturalist2017}
&  77.27  &  77.71
 & train & val[{\tiny:50\%}] & val[{\tiny50\%:}] & 
 \citep{Horn2018TheIS}
 \\
\href{https://www.tensorflow.org/datasets/catalog/i_naturalist2018}{i\_naturalist2018}
&  80.90  &  80.97
 & train & val[{\tiny:50\%}] & val[{\tiny50\%:}] & 
 \citep{Horn2018TheIS}
 \\
\href{https://www.tensorflow.org/datasets/catalog/imagenet_a}{imagenet\_a}
&  87.78  &  84.53
 & train[{\tiny10\%:}] & train[{\tiny5\%:10\%}] & train[{\tiny:5\%}] & 
 \citep{Hendrycks2021NaturalAE}
 \\
\href{https://www.tensorflow.org/datasets/catalog/imagenet_lt}{imagenet\_lt}
&  87.26  &  82.50
 & train & val & test & 
 \citep{Liu2019LargeScaleLR}
 \\
\href{https://www.tensorflow.org/datasets/catalog/imagenet_r}{imagenet\_r}
&  91.13  &  89.87
 & train[{\tiny10\%:}] & train[{\tiny5\%:10\%}] & train[{\tiny:5\%}] & 
 \citep{Hendrycks2021TheMF}
 \\
\href{https://www.tensorflow.org/datasets/catalog/imagenet_sketch}{imagenet\_sketch}
&  89.54  &  88.60
 & train[{\tiny10\%:}] & train[{\tiny5\%:10\%}] & train[{\tiny:5\%}] & 
 \citep{Wang2019LearningRG}
 \\
\href{https://www.tensorflow.org/datasets/catalog/imagenette}{imagenette}
&  99.92  &  100
 & train[{\tiny5\%:}] & val & train[{\tiny:5\%}] & 
 \citep{imagenette}
 \\
\href{https://www.tensorflow.org/datasets/catalog/imagewang}{imagewang}
&  97.32  &  99.59
 & train[{\tiny5\%:}] & val & train[{\tiny:5\%}] & 
 \citep{imagewang}
 \\
\href{https://www.tensorflow.org/datasets/catalog/malaria}{malaria}
&  98.17  &  97.46
 & train[{\tiny10\%:}] & train[{\tiny5\%:10\%}] & train[{\tiny:5\%}] &
 \citep{Rajaraman2018PretrainedCN}
 \\
\href{https://www.tensorflow.org/datasets/catalog/pet_finder}{pet\_finder}
&  62.40  &  60.73
 & train[{\tiny10\%:}] & train[{\tiny5\%:10\%}] & train[{\tiny:5\%}] &
 $-$ 
 \\
\href{https://www.tensorflow.org/datasets/catalog/places365_small}{places365\_small}
&  58.99  &  59.15
 & train & val[{\tiny:50\%}] & val[{\tiny50\%:}] &
 \citep{Zhou2018PlacesA1}
 \\
\href{https://www.tensorflow.org/datasets/catalog/plant_village}{plant\_village}
&  100  &  99.89
 & train[{\tiny10\%:}] & train[{\tiny5\%:10\%}] & train[{\tiny:5\%}] &
 \citep{Hughes2015AnOA}
 \\
\href{https://www.tensorflow.org/datasets/catalog/plantae_k}{plantae\_k}
&  99.05  &  90.74
 & train[{\tiny10\%:}] & train[{\tiny5\%:10\%}] & train[{\tiny:5\%}] &
 \citep{Kour2019PlantaeKAL}
 \\
\href{https://www.tensorflow.org/datasets/catalog/quickdraw_bitmap}{quickdraw\_bitmap}
&  78.35  &  77.59
 & train[{\tiny20k:}] & train[{\tiny10k:20k}] & train[{\tiny:10k}] & 
 \citep{Ha2018ANR}
 \\
\href{https://www.tensorflow.org/datasets/catalog/rock_paper_scissors}{rock\_paper\_scissors}
&  100  &  97.04
 & train[{\tiny5\%:}] & train[{\tiny:5\%}] & test &
 \citep{rps}
 \\
\href{https://www.tensorflow.org/datasets/catalog/siscore#rotation}{siscore/rotation}
&  100  &  100
 & train[{\tiny10\%:}] & train[{\tiny5\%:10\%}] & train[{\tiny:5\%}] &
 \citep{Djolonga2021OnRA}
 \\
\href{https://www.tensorflow.org/datasets/catalog/siscore#size}{siscore/size}
&  99.93  &  99.94
 & train[{\tiny10\%:}] & train[{\tiny5\%:10\%}] & train[{\tiny:5\%}] &
 \citep{Djolonga2021OnRA}
 \\
\href{https://www.tensorflow.org/datasets/catalog/siscore#location}{siscore/location}
&  99.99  &  99.95
 & train[{\tiny10\%:}] & train[{\tiny5\%:10\%}] & train[{\tiny:5\%}] &
 \citep{Djolonga2021OnRA}
 \\
\href{https://www.tensorflow.org/datasets/catalog/stanford_dogs}{stanford\_dogs}
&  95.17  &  93.50
 & train[{\tiny5\%:}] & train[{\tiny:5\%}] & test &
 \citep{Khosla2012NovelDF}
 \\
\href{https://www.tensorflow.org/datasets/catalog/stanford_online_products}{stanford\_online\_products} 
&  90.00  &  89.47
 & train & test[{\tiny:10k}] & test[{\tiny10k:}] & 
 \citep{Song2016DeepML}
 \\
\href{https://www.tensorflow.org/datasets/catalog/stl10}{stl10}
&  100  &  99.64
 & train[{\tiny5\%:}] & train[{\tiny:5\%}] & test &
 \citep{Coates2011AnAO}
 \\
\href{https://www.tensorflow.org/datasets/catalog/tf_flowers}{tf\_flowers}
&  99.45  &  97.83
 & train[{\tiny10\%:}] & train[{\tiny5\%:10\%}] & train[{\tiny:5\%}] &
 $-$ 
 \\
\href{https://www.tensorflow.org/datasets/catalog/uc_merced}{uc\_merced}
&  100  &  100
 & train[{\tiny10\%:}] & train[{\tiny5\%:10\%}] & train[{\tiny:5\%}] &
 \citep{Yang2010BagofvisualwordsAS}
 \\

    \bottomrule
  \end{tabular}
\end{table*}

\clearpage

\begin{algorithm}
\caption{Pseudocode for one active task iteration}
\label{algo}
\begin{algorithmic}[1]
\State Active task: $t$
\State Set of all the models currently in the multitask system: $\mathcal{M}$
\State Active population: $\mathcal{A} \gets \{m\ |\  m \in \mathcal{M} \land m$\ trained on $t \}$ 
\For{$\#generations$}
    \For{$\#child\mbox{-}models$}
        \State \Comment{Sample parent model}
        \State Parent model: $p \gets $ \textbf{none}
        \For{Candidate parent model: $\hat{p} \in [ sorted_{score}(\mathcal{A}), sorted_{random}(\mathcal{M}\setminus\mathcal{A})]$}  
            \If{$0.5^{\#selections(\hat{p}, t)} > x \sim Uniform([0, 1])$}
                \State $p \gets \hat{p}$
                \State \textbf{break}
            \EndIf
        \EndFor
        \If{$p=$ \textbf{none}}
            \State $p \sim Uniform(\mathcal{A} \cup \mathcal{M})$
        \EndIf
        \State  \Comment{Sample child model}
        \State Set of mutations: $\Delta \gets \{make\mbox{-}trainable\mbox{-}head\}$
        \For{Candidate mutation: $\hat{\delta} \in possible\mbox{-}mutations(p)$}
            \If{$\mu(\hat{\delta}|p) > x \sim Uniform([0, 1]) $}
                \State $\Delta \gets \Delta \cup \{\hat{\delta}\}$ 
            \EndIf
        \EndFor
        \State Untrained child model: $c_0 \gets apply\mbox{-}mutations(p, \Delta)$

        \State  \Comment{Train child model}
        \State Retained child model: $c \gets$ \textbf{none}
        \For{$i \in [1, ...\ , \#train\mbox{-}cycles]$}
            \State $c_i \gets train(c_{i-1}, min(1\ epoch,\ \#samples\mbox{-}cap))$
            \If{$score(c_i) \geq max( \{score(c) | c \not= $ \textbf{none}$\} \cup\{score(p) | p $ trained on $ t\} \cup \{\mbox{-}\infty\} )$}
                \State $c \gets c^i$
            \EndIf
        \EndFor
        \If{$c \not= $ \textbf{none}}
            \State $\mathcal{A} \gets \mathcal{A} \cup \{c\}$
        \EndIf
    \EndFor
\EndFor
\State \Comment{Keep only the best model for $t$}
\State $\mathcal{M} \gets \{ \argmax_{m \in \mathcal{A}} score(m)\} \cup \{m\ |\ m \in \mathcal{M} \land m $ not trained on $ t\}$

\end{algorithmic}
\end{algorithm}

\end{document}